\documentclass[10pt,twocolumn,letterpaper]{article}

\usepackage[pagenumbers]{wacv} 
\definecolor{wacvblue}{rgb}{0.21,0.49,0.74}
\usepackage[pagebackref,breaklinks,colorlinks,allcolors=wacvblue]{hyperref}

\def\wacvPaperID{1} 
\def\confName{WACV}
\def\confYear{2026}

\usepackage{microtype}
\usepackage{lipsum}
\usepackage{graphicx}
\usepackage{comment}
\usepackage{multirow}
\usepackage{array}
\usepackage{adjustbox}
\usepackage{rotating}
\usepackage{xcolor}
\usepackage{caption}
\usepackage{subcaption}
\usepackage[acronym]{glossaries}
\newacronym{ai}{AI}{Artificial Intelligence}
\newacronym{gdpr}{GDPR}{General Data Protection Regulation}
\newacronym{isic}{ISIC}{International Skin Imaging Collaboration}
\newacronym{pii}{PII}{Personal Identifiable Information}
\newacronym{tpr}{TPR}{True Positive Rate}
\newacronym{dp}{DP}{Demographic Parity}
\newacronym{genai}{GenAI}{Generative AI}
\newacronym{ham}{HAM}{HAM10000}
\newacronym{sota}{SOTA}{state-of-the-art}
\glsdisablehyper

\usepackage[margin=1in]{geometry}

\newcommand{\sampleimg}[3]{%
  \IfFileExists{imgs/synth_imgs_cfg4_steps400/#1_#2_#3.jpg}{%
    \includegraphics[
      width=18mm,
      height=18mm
    ]{imgs/synth_imgs_cfg4_steps400/#1_#2_#3.jpg}%
  }{%
    \fbox{\parbox{18mm}{\centering\vspace{18mm}}}%
  }%
}

\newcommand{\colheading}[2]{%
  \rotatebox{60}{\parbox{22mm}{\centering #1\\#2}}%
}

\newcommand{\rowlabel}[1]{\raisebox{3\height}{#1}}

\title{Towards Facilitated Fairness Assessment of AI-based Skin Lesion Classifiers Through GenAI-based Image Synthesis}

\author{Ko Watanabe$^*$\\
{\tt\small DFKI GmbH} \\
{\tt\small ko.watanabe@dfki.de}
\and
Stanislav Frolov$^*$\\
{\tt\small DFKI GmbH} \\
{\tt\small stanislav.frolov@dfki.de}
\and
Aya Hassan\\
{\tt\small RPTU Kaiserslautern-Landau} \\
{\tt\small aya.hassan@dfki.de}
\and
David Dembinsky\\
{\tt\small DFKI GmbH} \\
{\tt\small david.dembinsky@dfki.de}
\and
Adriano Lucieri\\
{\tt\small DFKI GmbH} \\
{\tt\small adriano.lucieri@dfki.de}
\\
\vspace{0.1em}
\footnotesize{$^*$Equal contribution}
\and
Andreas Dengel\\
{\tt\small DFKI GmbH} \\
{\tt\small andreas.dengel@dfki.de}
}

\begin{document}
\maketitle

\begin{abstract}
Recent advances in deep learning and on-device inference could transform routine screening for skin cancers. 
Along with the anticipated benefits of this technology, potential dangers arise from unforeseen and inherent biases. 
A significant obstacle is building evaluation datasets that accurately reflect key demographics, including sex, age, and race, as well as other underrepresented groups. 
To address this, we train a state-of-the-art generative model to generate synthetic data in a controllable manner to assess the fairness of publicly available skin cancer classifiers. 
To evaluate whether synthetic images can be used as a fairness testing dataset, we prepare a real-image dataset (MILK10K) as a benchmark and compare the True Positive Rate result of three models (DeepGuide, MelaNet, and SkinLesionDensnet).
As a result, the classification tendencies observed in each model when tested on real and generated images showed similar patterns across different attribute data sets.
We confirm that highly realistic synthetic images facilitate model fairness verification.
\end{abstract}

\section{Introduction}
Melanoma is the deadliest form of skin cancer, with an estimated 325,000 new cases (174,000 men and 151,000 women) and 57,000 deaths (32,000 men and 25,000 women) reported worldwide in 2020 \cite{arnold2022global}.
By 2040, incidence is projected to increase by nearly 50\% to 510,000 new cases, and mortality by 68\% to 96,000 deaths.
Early detection, however, dramatically improves survival.
For example, a population-wide screening project in Schleswig-Holstein, Germany, reported a reduction in melanoma mortality of 47\% for men and 49\% for women \cite{katalinic2012does}.
Similarly, screening efforts in Australia decreased melanoma deaths by 59\% \cite{watts2021association}.
These findings underline a crucial insight: despite rising incidence, early and accessible screening can substantially reduce mortality.

Recent advances in \gls{ai}, together with the widespread availability of high-resolution smartphone cameras, have created new opportunities for accessible melanoma pre-screening outside clinical environments.
\gls{ai}-assisted diagnostic systems can support clinicians during routine examinations and also provide human-understandable explanations for their decisions \cite{lucieri2022exaid}.
In principle, such systems could even enable remote pre-screening, allowing individuals to capture dermoscopic images themselves and receive preliminary risk assessments \cite{chao2017smartphone, rat2018use}.
These innovations have considerable societal potential, including earlier detection, reduced diagnostic delays, lower healthcare costs, and decreased clinician workload.

\begin{figure}[t]
  \centering
  \includegraphics[width=\columnwidth]{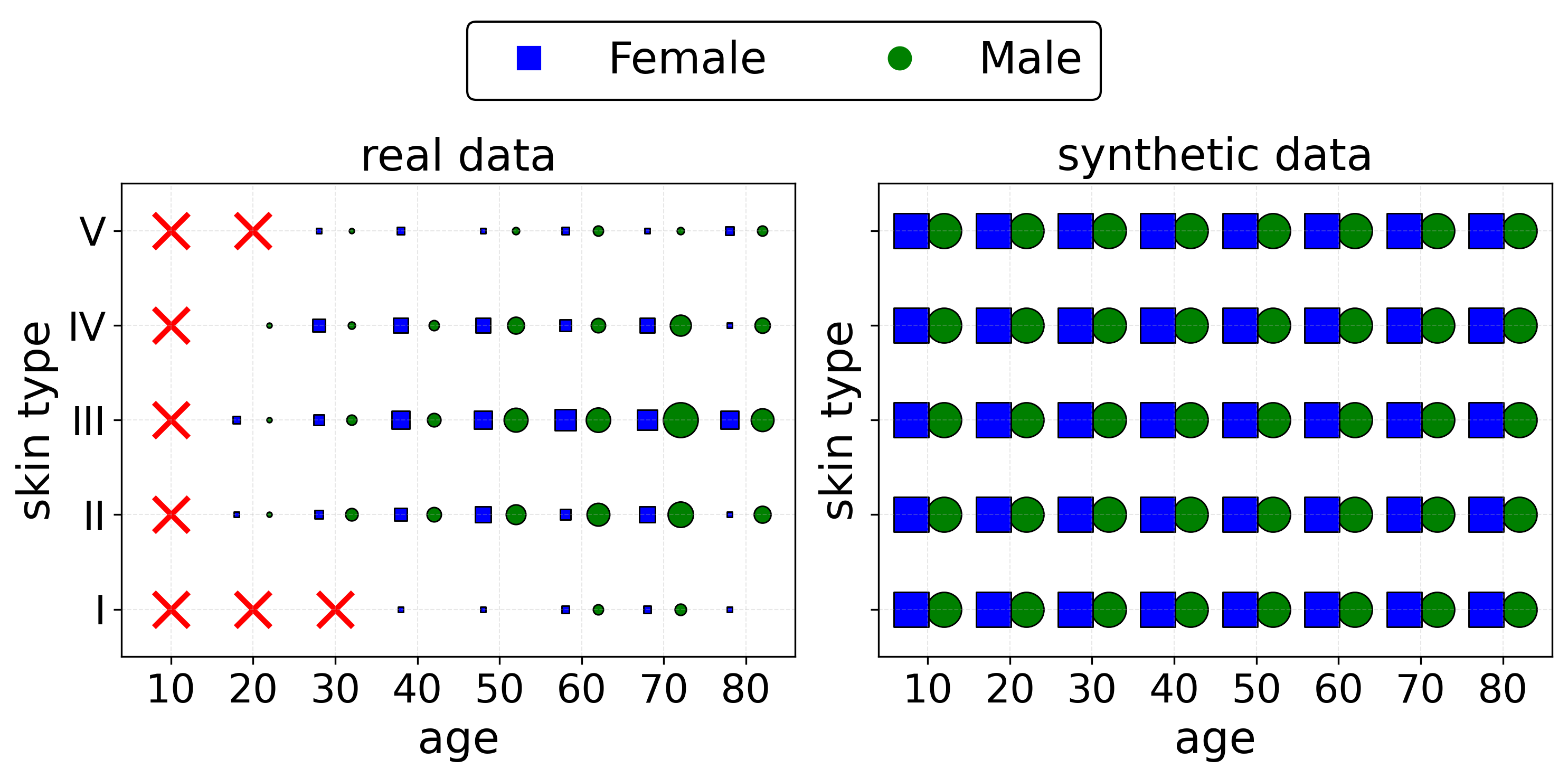}
  \caption{Real melanoma test data (left) such as MILK10k \cite{MILK10k_2025} shows limited and imbalanced demographic coverage, while synthetic data (right) can provide complete combinations with balanced sampling, enabling more reliable fairness assessment.}
  \label{fig:teaser}
\end{figure}

At the same time, the deployment of \gls{ai} in medicine requires a high level of trustworthiness, especially in light of regulatory developments such as the EU AI Act \cite{european2019ethics}.
Fairness and transparency in both datasets and trained models are essential conditions for reliable clinical use \cite{bevan2022detecting, ansari2024algorithmic, aayushman2024fair, bie2024xcoop, pahde2023reveal, dembinsky2025unifying}.
Fairness assessments depend on access to demographically well-characterized data. The \gls{isic} archive \cite{gutman2016skin} provides one of the most comprehensive publicly available skin lesion datasets \cite{gutman2016skin, codella2018skin, tschandl2018ham10000, codella2019skin, rotemberg2021patient, hernandez2024bcn20000}, including metadata such as sex, age, and Fitzpatrick skin type \cite{sachdeva2009fitzpatrick}.
Despite its scale and quality, demographic imbalance remains a major limitation and complicates fairness evaluations.

\begin{table*}[t!]
\centering
\small
\begin{tabular}{lllcl}
\toprule
\textbf{Model}                                 & \textbf{Architecture} & \textbf{Dataset}                      & \textbf{Public Weights} & \textbf{Performance} \\
\midrule
DeepGuide~(\cite{mallya2022deep})                  & DenseNet              & Derm7pt                               & \checkmark               & 0.788 (AUC)          \\
MelaNet~(\cite{zunair2020melanoma})                & VGG-GAP               & ISIC 2016                             & \checkmark               & 0.811 (AUROC)        \\
Patch-Lesion~(\cite{gessert2020skin})              & Dense121              & HAM10000                              & --                       & 0.855 (F1-Score)     \\
SkinLesionDensenet~(\cite{panda2021comparative})   & DenseNet201           & ISIC                                  & \checkmark               & 0.878 (F1-Score)     \\
EFFNet~(\cite{ma2023effnet})                       & EfficientNetV2        & HAM10000                              & --                       & 0.932 (F1-Score)     \\
\citet{sabir2024classification}                    & EfficientNet-B0       & HAM10000                              & --                       & 0.970 (F1-Score)     \\
\bottomrule
\end{tabular}
\caption{Comparison of existing melanoma detection models.}
\label{tab:melanoma_models_comparison}
\end{table*}

Meanwhile, recent progress in \gls{genai}, particularly diffusion models \cite{ho2020denoising, rombach2022high}, has enabled high-fidelity and attribute-controlled image synthesis.
In this work, we explore whether these advances can support fairness evaluation in melanoma detection.
Using the ISIC dataset and LightningDiT \cite{yao2025reconstruction}, a latent diffusion transformer \cite{peebles2023scalable}, we train a generative model capable of synthesizing realistic dermoscopic images conditioned on key patient attributes.
We then examine whether such synthetic cohorts can serve as a reliable resource for assessing algorithmic fairness in publicly available melanoma classifiers.
In summary, our work makes the following key contributions.
\begin{itemize}
    \item We present a diffusion-based framework for generating demographically balanced cohorts of dermoscopic images.
    Our model synthesizes high-fidelity skin lesion images conditioned on sex, age, and Fitzpatrick skin type.
    \item We introduce a protocol for fairness evaluation that applies the generated images to three peer-reviewed melanoma classifiers and quantifies true positive rate disparity.
    \item We provide an empirical analysis of the suitability of synthetic, demographically balanced cohorts for fairness auditing and discuss implications for future medical AI evaluation workflows.
\end{itemize}

\section{Related Work}
\label{sec:related_work}
The following sections review related work in three areas: generative models for synthetic data generation, melanoma detection models, and fairness evaluation in skin lesion classification.

\subsection{Generative Models for Skin Lesions}
Synthetic data generation is widely used to address data scarcity, class imbalance, and privacy concerns in skin lesion datasets.
Early approaches based on Generative Adversarial Networks (GANs) \cite{goodfellow2020generative}, such as MelanoGANs \cite{baur2018melanogans}, DCGAN \cite{radford2015unsupervised}, and LAPGAN \cite{denton2015deep}, demonstrated that realistic dermoscopic images can be synthesized even from limited data.
DermGAN \cite{ghorbani2019dermgan} and StyleGAN-based methods \cite{bissoto2019skin, karras2019style} introduced finer control over lesion characteristics and improved downstream classifier performance.
Later work, including StyleGAN2 \cite{karras2020training} and federated setups \cite{carrasco2022assessing}, further improved stability, fidelity, and privacy.
Diffusion models now dominate high-fidelity medical image synthesis.
Stable Diffusion \cite{rombach2022high} and DreamBooth \cite{ruiz2023dreambooth}, as applied in Cancer-Net SCa-Synth \cite{tai2024cancer}, support few-shot personalization and balanced dataset construction.
Extensions of latent diffusion models have also improved control over skin type variation and helped address fairness gaps \cite{wang2024majority}.
Our work differs by focusing on fairness evaluation rather than augmentation.
We train a state-of-the-art latent diffusion transformer based on LightningDiT \cite{yao2025reconstruction} to generate attribute-controlled dermoscopic image cohorts.
The model leverages all available \gls{isic} metadata and allows systematic control over demographic and clinical attributes for constructing balanced intersectional test sets.

\subsection{Melanoma Detection Models}
Prior research on melanoma classification has employed a variety of datasets, including ISIC, Derm7pt \cite{kawahara2018seven}, and HAM \cite{tschandl2018ham10000}, the latter containing 10,015 dermoscopic images covering seven lesion types. These datasets differ in composition, imaging style, and demographic distribution, which complicates direct model comparison.
A further challenge arises from the heterogeneous evaluation metrics reported in the literature.
Some studies rely on the F1-score, whereas others report AUC or AUROC, making cross-model comparison nontrivial. Pretrained model availability also varies.
Public weights exist for DeepGuide \cite{mallya2022deep}, MelaNet \cite{zunair2020melanoma}, and SkinLesionDensenet \cite{panda2021comparative}, but not for Patch-Lesion \cite{gessert2020skin} or EFFNet \cite{ma2023effnet}.
In this work, we use publicly available pretrained models to ensure reproducibility and consistent evaluation.

\subsection{Fairness in Skin Lesion Classification}
Fairness evaluation has gained increasing attention due to persistent performance disparities across demographic subgroups.
In \cite{ansari2024algorithmic}, the authors proposed an augmentation strategy that combines a source skin tone image with a diagnostic image to mitigate imbalance and skin tone bias.
Other approaches incorporate fairness directly into model design.
BiaslessNAS \cite{sheng2024data} integrates fairness constraints throughout the neural architecture search process and reports improvements of 2.55\% in accuracy and 65.50\% in fairness over baseline models.
Compared to previous augmentation-based fairness approaches, our method generates attribute-controlled synthetic images through text prompts.
We evaluate fairness in existing pretrained melanoma classifiers \cite{mallya2022deep, zunair2020melanoma, panda2021comparative} using balanced synthetic cohorts, allowing us to investigate the potential of prompt-driven generative models as a tool for systematic fairness assessment.

\section{Methodology}
\begin{figure*}[t!]
  \centering
  \includegraphics[width=0.9\textwidth]{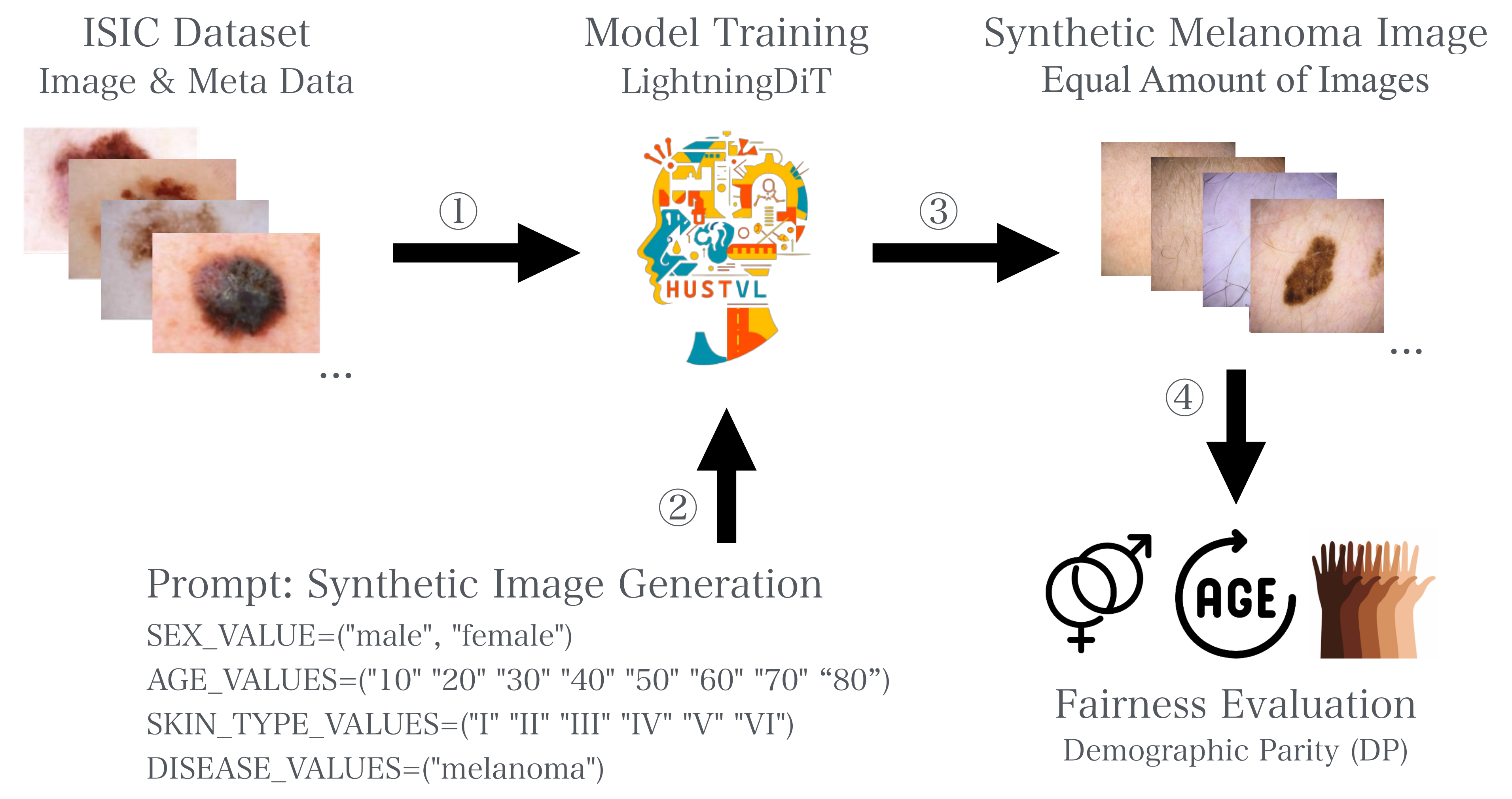}
  \caption{Overall pipeline of our fairness testing workflow. We first train a generative model based on LightningDiT \cite{yao2025reconstruction} using the \gls{isic} dataset. Synthetic melanoma test images are then generated via systematic prompt templates. Finally, the synthetic images are used to assess fairness as measured via \gls{tpr} difference in pre-trained melanoma detection models.}
  \label{fig:test_workflow}
\end{figure*}

In this section, we first describe the process for generating synthetic test data.
Next, we describe the fairness evaluation process against the selected pre-trained models for melanoma detection.
In \autoref{fig:test_workflow}, we illustrate our complete fairness testing workflow.

\subsection{Synthetic Test Data Generation}
\label{sec:synthetic_test_data}

\noindent
\textbf{Model:}
Our fairness evaluation relies on a controllable generative model capable of producing high-fidelity dermoscopic images across clinically and demographically relevant attributes. We train a latent diffusion transformer using the LightningDiT framework \cite{yao2025reconstruction}. LightningDiT improves reconstruction and generative quality by aligning the underlying variational autoencoder (VAE) \cite{kingma2013auto} with pretrained vision foundation models. The resulting vision foundation model aligned VAE (VA-VAE) produces latent representations that are consistent with models such as DINOv2 \cite{oquab2023dinov2} and MAE \cite{he2022masked}, which enhances semantic structure and supports reliable attribute conditioning. The latent diffusion transformer contains 675M parameters and is trained for 80,000 steps using AdamW \cite{loshchilov2019decoupled}, a learning rate of 0.0002, mixed precision, and a global batch size of 1024 on 8$\times$A100 GPUs. Training requires approximately 8 hours.

\noindent
\textbf{Data:}
Only publicly available data are used and no new patient information is collected. All dermoscopic images from the \gls{isic} archive (2016--2024) are processed at a resolution of $256 \times 256$ and encoded into the latent space using the VA-VAE, resulting in roughly 500{,}000 latent-image pairs. Each image is paired with a text description constructed from the available annotations, including sex, age group, lesion size, Fitzpatrick skin type, and diagnosis. Data with insufficient attributes are excluded. Text embeddings are produced using CLIP \cite{radford2021learning}, a contrastively trained vision-language model that embeds images and text into a shared latent space and provides strong semantic grounding for conditioning.

\noindent
\textbf{Generation:}
After training, the model acts as a controllable generator for constructing balanced synthetic test sets. Synthetic dermoscopic images are produced by sampling random latent noise and conditioning the model on attribute-specific text prompts that target demographic groups under evaluation. We use the Euler solver with 400 diffusion steps and a classifier-free guidance scale of 4. The resulting images form consistent, attribute-controlled cohorts for downstream fairness assessment.

\subsection{Balanced Test Data Generation}
\label{subsec:data_balance}

\noindent
\textbf{Overview:}
To evaluate fairness in melanoma classification models, we create a synthetic test set that is explicitly balanced across relevant \gls{pii} attributes. As illustrated in \autoref{fig:test_workflow}, images are generated through a systematic prompting procedure applied to our trained generative model. Each prompt specifies the demographic attributes of interest while other visual factors are held as consistent as possible. This controlled setup reduces confounding variation and ensures that differences in classifier predictions can be traced to subgroup-specific effects rather than distributional imbalance.

\vspace{0.35em}
\noindent
\textbf{Prompt Design:}
We adopt a full-factorial scheme across four \gls{pii} dimensions: sex (2 categories), age group (8 categories), Fitzpatrick skin type (6 categories), and disease case (1 category corresponding to melanoma). This yields 96 unique attribute configurations ($2 \times 8 \times 6 \times 1$), covering key demographic and phenotypic factors known to influence dermatological diagnosis.

\vspace{0.25em}
\noindent
\textbf{Sampling Strategy:}
For each configuration, we generate 850 synthetic melanoma images to ensure equal subgroup representation. The resulting test set contains 81,600 images and provides a balanced distribution across all selected \gls{pii} factors. This design enables a controlled and statistically robust assessment of fairness in pretrained melanoma classification models. 

\begin{figure*}[t]
\centering
\vspace{1em}

\resizebox{\textwidth}{!}{%
\setlength{\tabcolsep}{2pt} 

\begin{tabular}{
    c
    *{2}{c}@{\hspace{2pt}}  
    *{2}{c}@{\hspace{2pt}}  
    *{2}{c}@{\hspace{2pt}}  
    *{2}{c}@{\hspace{2pt}}  
    *{2}{c}@{\hspace{2pt}}  
    *{2}{c}                 
}
\rowlabel{Age 10} &
\sampleimg{10}{female}{I} & \sampleimg{10}{male}{I} &
\sampleimg{10}{female}{II} & \sampleimg{10}{male}{II} &
\sampleimg{10}{female}{III} & \sampleimg{10}{male}{III} &
\sampleimg{10}{female}{IV} & \sampleimg{10}{male}{IV} &
\sampleimg{10}{female}{V} & \sampleimg{10}{male}{V} &
\sampleimg{10}{female}{VI} & \sampleimg{10}{male}{VI} \\

\rowlabel{Age 20} &
\sampleimg{20}{female}{I} & \sampleimg{20}{male}{I} &
\sampleimg{20}{female}{II} & \sampleimg{20}{male}{II} &
\sampleimg{20}{female}{III} & \sampleimg{20}{male}{III} &
\sampleimg{20}{female}{IV} & \sampleimg{20}{male}{IV} &
\sampleimg{20}{female}{V} & \sampleimg{20}{male}{V} &
\sampleimg{20}{female}{VI} & \sampleimg{20}{male}{VI} \\

\rowlabel{Age 30} &
\sampleimg{30}{female}{I} & \sampleimg{30}{male}{I} &
\sampleimg{30}{female}{II} & \sampleimg{30}{male}{II} &
\sampleimg{30}{female}{III} & \sampleimg{30}{male}{III} &
\sampleimg{30}{female}{IV} & \sampleimg{30}{male}{IV} &
\sampleimg{30}{female}{V} & \sampleimg{30}{male}{V} &
\sampleimg{30}{female}{VI} & \sampleimg{30}{male}{VI} \\

\rowlabel{Age 40} &
\sampleimg{40}{female}{I} & \sampleimg{40}{male}{I} &
\sampleimg{40}{female}{II} & \sampleimg{40}{male}{II} &
\sampleimg{40}{female}{III} & \sampleimg{40}{male}{III} &
\sampleimg{40}{female}{IV} & \sampleimg{40}{male}{IV} &
\sampleimg{40}{female}{V} & \sampleimg{40}{male}{V} &
\sampleimg{40}{female}{VI} & \sampleimg{40}{male}{VI} \\

\rowlabel{Age 50} &
\sampleimg{50}{female}{I} & \sampleimg{50}{male}{I} &
\sampleimg{50}{female}{II} & \sampleimg{50}{male}{II} &
\sampleimg{50}{female}{III} & \sampleimg{50}{male}{III} &
\sampleimg{50}{female}{IV} & \sampleimg{50}{male}{IV} &
\sampleimg{50}{female}{V} & \sampleimg{50}{male}{V} &
\sampleimg{50}{female}{VI} & \sampleimg{50}{male}{VI} \\

\rowlabel{Age 60} &
\sampleimg{60}{female}{I} & \sampleimg{60}{male}{I} &
\sampleimg{60}{female}{II} & \sampleimg{60}{male}{II} &
\sampleimg{60}{female}{III} & \sampleimg{60}{male}{III} &
\sampleimg{60}{female}{IV} & \sampleimg{60}{male}{IV} &
\sampleimg{60}{female}{V} & \sampleimg{60}{male}{V} &
\sampleimg{60}{female}{VI} & \sampleimg{60}{male}{VI} \\

\rowlabel{Age 70} &
\sampleimg{70}{female}{I} & \sampleimg{70}{male}{I} &
\sampleimg{70}{female}{II} & \sampleimg{70}{male}{II} &
\sampleimg{70}{female}{III} & \sampleimg{70}{male}{III} &
\sampleimg{70}{female}{IV} & \sampleimg{70}{male}{IV} &
\sampleimg{70}{female}{V} & \sampleimg{70}{male}{V} &
\sampleimg{70}{female}{VI} & \sampleimg{70}{male}{VI} \\

\rowlabel{Age 80} &
\sampleimg{80}{female}{I} & \sampleimg{80}{male}{I} &
\sampleimg{80}{female}{II} & \sampleimg{80}{male}{II} &
\sampleimg{80}{female}{III} & \sampleimg{80}{male}{III} &
\sampleimg{80}{female}{IV} & \sampleimg{80}{male}{IV} &
\sampleimg{80}{female}{V} & \sampleimg{80}{male}{V} &
\sampleimg{80}{female}{VI} & \sampleimg{80}{male}{VI} \\

&
\colheading{Skin Type I}{Female} &
\colheading{Skin Type I}{Male} &
\colheading{Skin Type II}{Female} &
\colheading{Skin Type II}{Male} &
\colheading{Skin Type III}{Female} &
\colheading{Skin Type III}{Male} &
\colheading{Skin Type IV}{Female} &
\colheading{Skin Type IV}{Male} &
\colheading{Skin Type V}{Female} &
\colheading{Skin Type V}{Male} &
\colheading{Skin Type VI}{Female} &
\colheading{Skin Type VI}{Male} \\

\end{tabular}
} 
\caption{Synthetic melanoma images generated by our model. Rows represent Fitzpatrick skin types (I--VI) combined with sex, and columns represent age groups (10--80). The grid demonstrates coverage of diverse demographic groups for fairness assessment.}
  \label{fig:melanoma_samples}
\end{figure*}

\subsection{Fairness Assessment}
\label{subsec:fairness}

\noindent
\textbf{Metric Definition:}
Since our synthetic test set contains only melanoma (\ie positive) cases, we quantify subgroup disparities using \gls{tpr} disparity~\cite{hardt2016equality}, defined as the difference between the maximum and minimum true positive rates across \gls{pii} groups.
In a positive-only setting, the proportion of predicted positives in each group coincides with sensitivity. For a single decision threshold~$\tau$ shared across all groups:

\begin{align}
\mathrm{TPR}(g) &= \frac{1}{\lvert D_g^{+}\rvert} \sum_{(x,y)\in D_g^{+}} \mathbf{1}\{\hat{f}(x)\ge \tau\}, \label{eq:tpr-definition} \\
\Delta_{\mathrm{TPR}} &= \max_{g}\,\mathrm{TPR}(g) - \min_{g}\,\mathrm{TPR}(g)
\label{eq:tpr-disparity}
\end{align}
We compute $\Delta_{\mathrm{TPR}}$ on synthetic cohorts stratified by sex, age group, and Fitzpatrick skin type.
To generate a strictly positive-only test set, we use ``melanoma'' as the disease value in our systematic prompt configuration.
The resulting measure, $\Delta_{\mathrm{TPR}}$, therefore provides a meaningful and statistically robust indicator of whether a model systematically favors or disadvantages particular demographic groups~\cite{stanley2022disproportionate, li2021estimating, kinyanjui2020fairness}.


\noindent
\textbf{Evaluation Protocol:}
The balanced and fully factorial nature of the synthetic test set enables unbiased comparisons across demographic subgroups. Although the dataset contains only positive cases for computing \gls{tpr}, it can also be used for additional performance metrics whose thresholds are determined externally, such as sensitivity, specificity, and AUROC. This setup supports both macro-averaged and worst-case subgroup analysis, and thus provides a rigorous basis for evaluating fairness in melanoma classification models.

\noindent
\textbf{Models:}
We evaluate three publicly available pretrained melanoma classifiers: DeepGuide~\cite{mallya2022deep}, MelaNet \cite{zunair2020melanoma}, and SkinLesionDensenet~\cite{panda2021comparative}. 
These models were selected due to the availability of pretrained weights and their relevance in prior benchmark studies. \textit{MelaNet} and \textit{SkinLesionDensenet} are trained on \gls{isic}, whereas \textit{DeepGuide} is trained on the \gls{ham} dataset, providing complementary training distributions. Architecturally, \textit{DeepGuide} and \textit{SkinLesionDensenet} rely on DenseNet backbones, while \textit{MelaNet} is based on a VGG-GAP design. This diversity in both training data and network architecture offers a representative foundation for examining fairness behavior across classifier families.

\begin{figure*}[t!]
  \centering
  \begin{subfigure}[t]{0.9\textwidth}
    \centering
    \includegraphics[width=\textwidth]{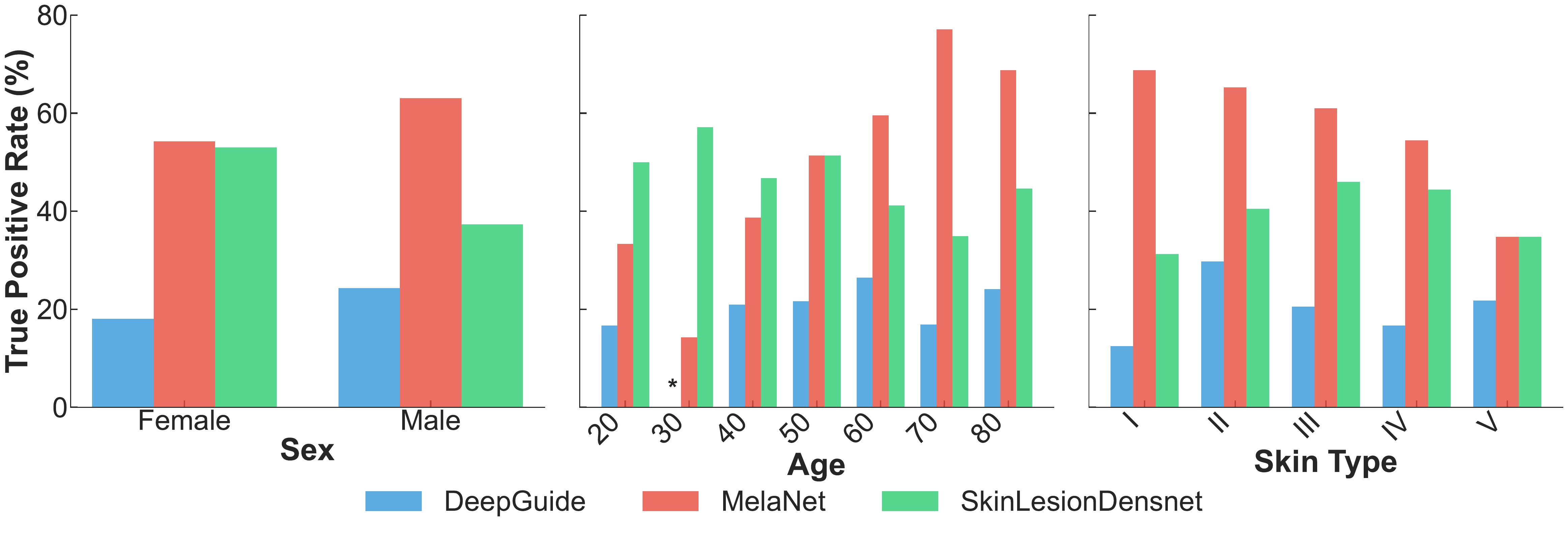}
    \caption{Real images from MILK10K}
    \label{fig:fairness_evaluation_sex}
  \end{subfigure}
  \hfill
  \begin{subfigure}[t]{0.9\textwidth}
    \centering
    \includegraphics[width=\textwidth]{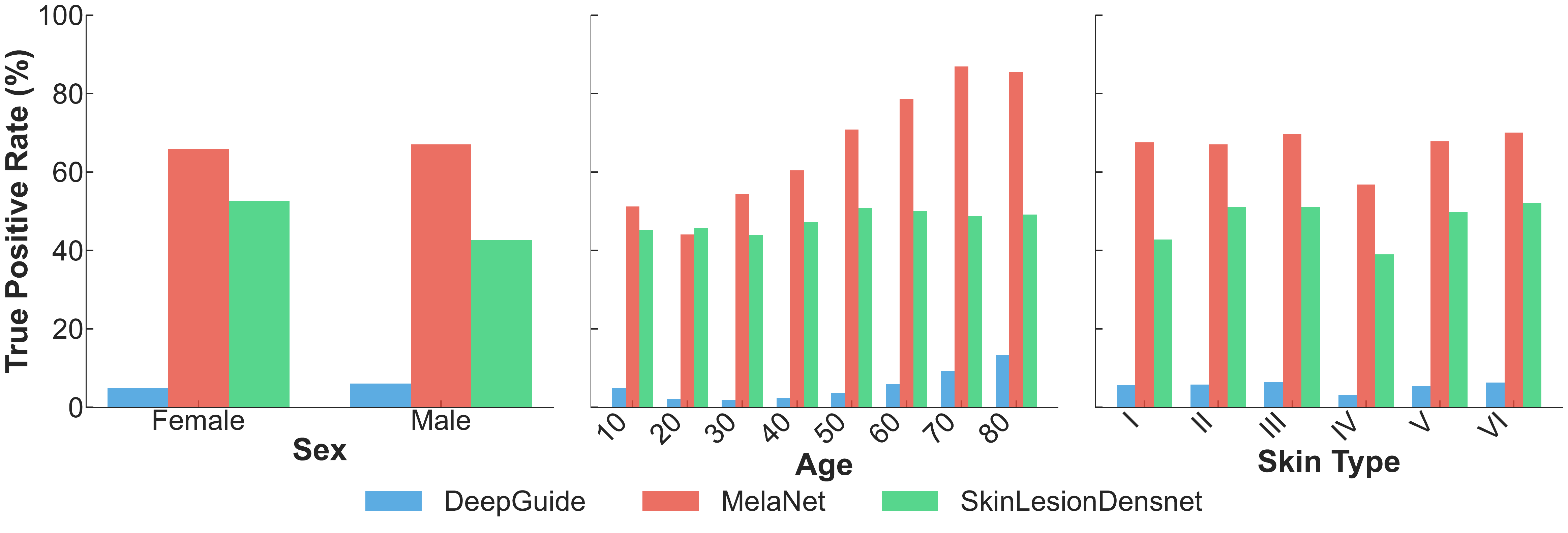}
    \caption{Synthetic images generated by our method}
    \label{fig:fairness_evaluation_age}
  \end{subfigure}
  \caption{\gls{tpr} of the skin lesion classifier across different \gls{pii} (sex, age, and Fitzpatrick skin type) groups. The result present that both real and synthetic image testing perform a similar bias trend for each \textit{DeepGuide}, \textit{MelaNet}, and \textit{SkinLesionDensnet}.}
  \label{fig:fairness_evaluation}
\end{figure*}

\begin{figure*}[t!]
  \centering
  \includegraphics[width=\textwidth]{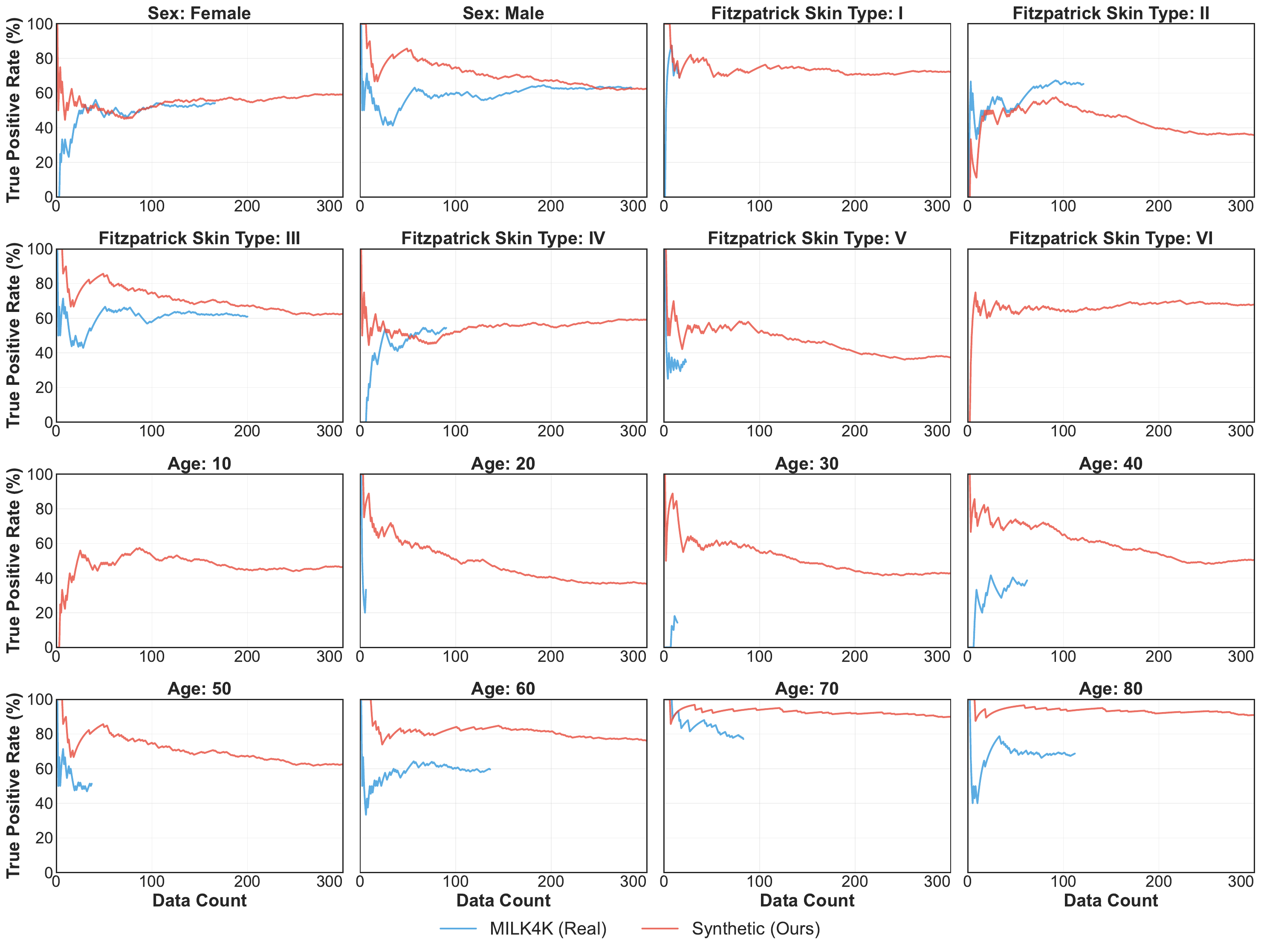}
  \caption{Cumulative \gls{tpr} curves for the PII attributes on MILK10K (real) and our synthetic images using MelaNet. The results show that the \gls{tpr} converges to similar levels, particularly for sex and skin type, indicating that synthetic images can serve as a suitable fairness evaluation dataset.}
  \label{fig:data_count_vs_accuracy}
\end{figure*}

\section{Results and Discussion}

\subsection{Generated Synthetic Skin Lesion Images}
\autoref{fig:melanoma_samples} presents representative melanoma images synthesized by our model under controlled prompting. The vertical axis enumerates Fitzpatrick skin types (I--VI) for each sex (male and female), while the horizontal axis spans age categories in 10-year increments from 10 to 80 years. This grid structure facilitates visual inspection of demographic variation. The generator responds to conditioning in a largely coherent manner, with clear subgroup-specific differences in appearance.

While the generated images are highly realistic, one limitation is immediately noticeable.
The generated images do not became increasingly darker with higher Fitzpatrick skin type.
For example, the generated images do not become progressively darker with higher Fitzpatrick skin types. Although this might be the intuitive expectation, prior work \cite{tong2020investigating,kalb2023revisiting} shows that lighting, camera settings, and lesion visibility introduce substantial variability, making pixel intensity an unreliable indicator. Further details and visual comparisons with real data exhibiting the same issue are provided in the appendix.

Given that downstream fairness analysis relies critically on the quality and diversity of the synthetic images, principled quality-control procedures are essential. Section~\ref{subsec:quality_control} discusses possible approaches, including attribute-specific judging models and screening using vision–language models prior to classifier evaluation. Such methods are intended to detect unrealistic or low-diversity generations before they influence fairness conclusions.

\subsection{Evaluation of Fairness Performance}

\noindent
\textbf{Group-wise TPR on Real and Synthetic Cohorts:}
\autoref{fig:fairness_evaluation} summarizes group-wise fairness results for the three pretrained melanoma classifiers across sex, age group, and Fitzpatrick skin type. Since the synthetic test set contains only melanoma (positive) cases, the melanoma prediction rate shown in the figure corresponds directly to sensitivity. To compare real and synthetic testing, we compute the performance on MILK4K and on the synthetic cohort separately. Following Section~\autoref{subsec:fairness}, we apply a single shared decision threshold and quantify disparity as the difference between the maximum and minimum \gls{tpr} values within each attribute.
Overall, real and synthetic image testing exhibit similar trends. When sex is treated as the target attribute, both \textit{DeepGuide} and \textit{MelaNet} achieve higher accuracy for male than for female, whereas with the same test images \textit{SkinLesionDensenet} performs better for female than for male. This demonstrates that synthetic data can reproduce the gender bias observed on real-world data. For age, \textit{MelaNet} shows that older age groups are associated with higher \gls{tpr} for both MILK4K and synthetic images, indicating that age-related biases are also reflected in the synthetic cohort. For skin type, the performance ranking of \textit{MelaNet}, \textit{SkinLesionDensenet}, and \textit{DeepGuide} is consistent between real and synthetic testing.
A plausible explanation for the skin-type trends is a distributional shift between training and test data. Both \textit{MelaNet} and \textit{SkinLesionDensenet} are trained on \gls{isic}, whereas \textit{DeepGuide} is trained on the \gls{ham} dataset. Because the training and test sources differ, similar relative performance patterns emerge across both real and synthetic image testing.

\noindent
\textbf{Effect of Test Set Size:}
\autoref{fig:data_count_vs_accuracy} shows cumulative \gls{tpr} as a function of the number of test images. The data count corresponds to the number of images used and illustrates how recognition accuracy changes as more test data are included. As a rule of thumb, as the number of test images increases, convergence to a stable accuracy is expected~\cite{rolnick2017deep}. Therefore, if accuracy trends for real and synthetic images converge, fairness simulations can be reliably performed using generated images.
For sex, both male and female curves exhibit this convergence behavior, and synthetic testing closely tracks real-image performance as the available data volume increases. In contrast, the age-related curves for real and synthetic data do not fully converge, suggesting that the generator and real data distributions differ more strongly along the age dimension. Age is a factor with substantial intra-group variability (e.g., early versus late decades), which may contribute to these discrepancies and should be examined in future work.

\noindent
\textbf{Attribute-wise TPR Disparities:}
\autoref{tab:dp_by_attribute} reports the maximum, minimum, and range of \gls{tpr} across subgroups for each attribute. For sex and skin type, \textit{MelaNet} achieves both the highest overall performance and the smallest disparity $\Delta_{\mathrm{TPR}}$. On the synthetic cohort, $\Delta_{\mathrm{TPR}}$ for sex is 0.0115 and for Fitzpatrick skin type is 0.1322. Interestingly, for the age attribute, \textit{MelaNet} exhibits the largest disparity, whereas \textit{SkinLesionDensenet} is substantially fairer: its $\Delta_{\mathrm{TPR}}$ for age is 0.0682, the lowest among the three models. These patterns are consistent between real and synthetic testing, reinforcing the usefulness of the synthetic cohort for fairness comparison.

\begin{table*}[t!]
\centering
\footnotesize
\setlength{\tabcolsep}{3pt}
\begin{tabular}{l l ccc ccc ccc}
\toprule
\textbf{Model} & \textbf{Dataset} &
\multicolumn{3}{c}{\textbf{Sex}} &
\multicolumn{3}{c}{\textbf{Age}} &
\multicolumn{3}{c}{\textbf{Fitzpatrik Skin Type}} \\
\cmidrule(lr){3-5}
\cmidrule(lr){6-8}
\cmidrule(lr){9-11}
 & &
\textbf{Max} & \textbf{Min} & $\Delta_{\mathrm{TPR}}$ $\downarrow$ &
\textbf{Max} & \textbf{Min} & $\Delta_{\mathrm{TPR}}$ $\downarrow$ &
\textbf{Max} & \textbf{Min} & $\Delta_{\mathrm{TPR}}$ $\downarrow$ \\
\midrule

DeepGuide & Ours
& 0.0601 (M) & 0.0487 (F) & 0.0114
& 0.1337 (80) & 0.0189 (30) & 0.1148
& 0.0643 (III) & 0.0311 (IV) & 0.0332 \\

 & MILK4K
& 0.2430 (M) & 0.1807 (F) & 0.0622
& 0.2647 (60) & 0.1667 (20) & 0.0980
& 0.2975 (II) & 0.1250 (I) & 0.1725 \\

\midrule
MelaNet & Ours
& 0.6706 (M) & 0.6591 (F) & 0.0115
& 0.8693 (70) & 0.4404 (20) & 0.4289
& 0.7004 (VI) & 0.5682 (IV) & 0.1322 \\

 & MILK4K
& 0.6303 (M) & 0.5422 (F) & 0.0881
& 0.7711 (70) & 0.1429 (30) & 0.6282
& 0.6875 (I) & 0.3478 (V) & 0.3397 \\

\midrule
SkinLesionDensenet & Ours
& 0.5255 (F) & 0.4267 (M) & 0.0988
& 0.5081 (50) & 0.4399 (30) & 0.0682
& 0.5209 (VI) & 0.3899 (IV) & 0.1310 \\

 & MILK4K
& 0.5301 (F) & 0.3732 (M) & 0.1569
& 0.5714 (30) & 0.3494 (70) & 0.2220
& 0.4600 (III) & 0.3125 (I) & 0.1475 \\

\bottomrule
\end{tabular}
\caption{Performance comparison of maximum (Max) \gls{tpr}, minimum (Min) \gls{tpr}, and \gls{tpr} difference ($\Delta_{\mathrm{TPR}}$) across attributes (Sex, Age, Skin Type). The symbol next to the value indicates which attribute it corresponds to, such as M is Male and F is Female.}
\label{tab:dp_by_attribute}
\end{table*}

\noindent
\textbf{Methodological Implications:}
These results illustrate two methodological advantages of our approach. First, attribute-conditioned synthetic data enable direct, fine-grained comparisons of subgroup sensitivities within a controlled, fully balanced design. Second, by training generators on different data sources, fairness analysis can be extended to examine the influence of the training data itself.
In practice, we recommend a two-step protocol: (i) verify that a pretrained model achieves reasonable sensitivity on synthetic images that match its training distribution, and only then (ii) assess subgroup disparities. This procedure yields more reliable conclusions regarding model fairness and highlights the importance of dataset–model alignment in fairness evaluations.

\section{Limitations and Future Work}
\label{sec:limitation}
In the following, we discuss limitations and future research directions.

\subsection{Enhancing Dataset Use for Training}
In this work, we generated synthetic image test data by training a custom LightningDiT \cite{yao2025reconstruction}.
According to \autoref{fig:melanoma_samples}, the images show high realism, and \autoref{fig:data_count_vs_accuracy} confirms a similar trend in the cumulative \gls{tpr} comparison between real and synthetic images.
However, according to \autoref{fig:data_count_vs_accuracy}, age \gls{tpr} conversion differs between real and synthetic images.
This may be due to the training procedure, which determines the precision of the label used for training.
The next step to improve the synthetic image generation model is to include a detail attribute in the training procedure of our generative model. 
Another direction is to add multiple datasets and use different image capturing devices to make the dataset and thus generation process more robust.

\subsection{Quality Control of the Synthetic Images}
\label{subsec:quality_control}
In our work, we evaluate fairness performance through 81,600 synthetic images generated in a pattern presented in \autoref{subsec:data_balance}.
To ensure the quality of the generated images, further quality control models could be used that evaluate the reliability of the images~\cite{pmlr-v136-kwon20a}.
The approach may, for example, be to implement quality-judging models for each age, sex, and skin color.
Also, due to the significant recent growth of visual language models (VLMs)~\cite{zhang2024vision}, VLM-as-a-judge may be an approach to perform quality evaluation before model prediction.
Hence, the next step in the current architecture is also to consider new models for quality assurance of the test images.

\subsection{Understanding Causes of Unfairness}
Our results verified that synthetic image generation can serve as test data for evaluating the fairness of pre-trained models.
For instance, when examining the performance of pre-trained models, we confirm that for sex and skin type, \textit{MelaNet} shows the most fair behavior, with $\Delta_{\mathrm{TPR}}$ of 0.0115 and 0.1322. For age \textit{SkinLesionDensnet} performs the best with 0.0682.
Although we have managed to measure $\Delta_{\mathrm{TPR}}$ for model fairness comparison, we still lack a method for discovering the causes or factors of unfairness.
Hence, the next step is to apply concept-based fairness~\cite{nejadgholi2022towards, pendyala2022concept, tong2020investigating} evaluation to understand the actual cause of the fairness issue.

\subsection{Various Diagnoses for Testing}
We focused on \gls{tpr} assessed through the disparity in true positive rates ($\Delta_{\mathrm{TPR}}$) under a melanoma-only (positive) test design. 
To broaden the scope of fairness evaluation, we aim to extend the generator and evaluation protocol to multiple diagnostic categories (e.g., benign nevi and other lesion types) and to multi-class classifiers in the future. 
This will enable (i) analysis of misclassification patterns (false positives and false negatives) by \gls{pii} group, (ii) computation of additional group fairness criteria beyond $\Delta_{\mathrm{TPR}}$, such as equalized odds~\cite{hardt2016equality}, and (iii) reporting of worst-group and macro-averaged metrics in the multi-class setting. 

\subsection{Synthetic Data for Facilitating Fair Models}
In this study, we verified whether synthetic data can be used for \gls{ai} model fairness testing.
We found that our generated and synthetic images exhibit a similar trend in fairness prediction.
The next step is to use synthetic images to train the model fairly by providing a balanced dataset.
Thus, we plan to use the synthetic images for fair model implementation in the future.

\section{Clinical Relevance Statement}
Our work addresses the risk that \gls{ai}‑based melanoma classifiers perform unevenly across sex, age, and skin type because current dermoscopic datasets are demographically imbalanced. In contrast, the current standard treatment involves examination by a dermatologist and dermoscopy testing, with teledermatology and simple \gls{ai} tools sometimes used as supplementary measures. Our controllable diffusion model generates demographically balanced synthetic melanoma cohorts and uses them to measure true‑positive‑rate disparities, enabling systematic, privacy‑preserving fairness audits that are difficult with real data alone. For clinical use, the generator must be trained on representative data, coupled with automated and expert image quality control, and validated prospectively to show that synthetic‑cohort fairness patterns predict disparities in real patients without reducing accuracy. Key risks include poor modeling of rare phenotypes, amplification of hidden data biases, distribution shifts between synthetic and clinical images, and over-reliance on synthetic audits in place of multi-center trials.

\section{Conclusion}
In this work, we trained a state-of-the-art generative model for controllable synthetic melanoma test image generation.
We verified that synthetic images can serve as test data for evaluating the fairness of pre-trained models.
We also found that performance declines modestly when the classifier and the generator are trained on disjoint datasets, underscoring the impact of dataset shift.
This means that a strictly privacy-preserving, \gls{pii}-free fairness audit is most reliable when both the generator and the detector originate from the same data distribution.
Beyond fairness assessment, the synthetic test set proved helpful for stress-testing model robustness, suggesting that data-driven generation can serve as a general diagnostic tool for \gls{ai} in medical imaging.

\subsection*{Acknowledgements}
Anonymized for the review process.

{
    \small
    \bibliographystyle{ieeenat_fullname}
    \bibliography{main}

@article{lucieri2022exaid,
  title     = {ExAID: A multimodal explanation framework for computer-aided diagnosis of skin lesions},
  author    = {Lucieri, Adriano and Bajwa, Muhammad Naseer and Braun, Stephan Alexander and Malik, Muhammad Imran and Dengel, Andreas and Ahmed, Sheraz},
  journal   = {Computer Methods and Programs in Biomedicine},
  volume    = {215},
  pages     = {106620},
  year      = {2022},
  publisher = {Elsevier}
}

@article{dembinsky2025unifying,
  title   = {Unifying VXAI: A Systematic Review and Framework for the Evaluation of Explainable AI},
  author  = {Dembinsky, David and Lucieri, Adriano and Frolov, Stanislav and Najjar, Hiba and Watanabe, Ko and Dengel, Andreas},
  journal = {arXiv preprint arXiv:2506.15408},
  year    = {2025}
}

@article{arnold2022global,
  title     = {Global burden of cutaneous melanoma in 2020 and projections to 2040},
  author    = {Arnold, Melina and Singh, Deependra and Laversanne, Mathieu and Vignat, Jerome and Vaccarella, Salvatore and Meheus, Filip and Cust, Anne E and De Vries, Esther and Whiteman, David C and Bray, Freddie},
  journal   = {JAMA dermatology},
  volume    = {158},
  number    = {5},
  pages     = {495--503},
  year      = {2022},
  publisher = {American Medical Association}
}

@article{katalinic2012does,
  title     = {Does skin cancer screening save lives? An observational study comparing trends in melanoma mortality in regions with and without screening},
  author    = {Katalinic, Alexander and Waldmann, Annika and Weinstock, Martin A and Geller, Alan C and Eisemann, Nora and Greinert, Ruediger and Volkmer, Beate and Breitbart, Eckhard},
  journal   = {Cancer},
  volume    = {118},
  number    = {21},
  pages     = {5395--5402},
  year      = {2012},
  publisher = {Wiley Online Library}
}

@article{kingma2013auto,
  title={Auto-encoding variational bayes},
  author={Kingma, Diederik P and Welling, Max},
  journal={arXiv preprint arXiv:1312.6114},
  year={2013}
}

@article{watts2021association,
  title     = {Association between melanoma detected during routine skin checks and mortality},
  author    = {Watts, Caroline G and McLoughlin, Kirstie and Goumas, Chris and Van Kemenade, Cathelijne H and Aitken, Joanne F and Soyer, H Peter and Pe{\~n}as, Pablo Fernandez and Guitera, Pascale and Scolyer, Richard A and Morton, Rachael L and others},
  journal   = {JAMA dermatology},
  volume    = {157},
  number    = {12},
  pages     = {1425--1436},
  year      = {2021},
  publisher = {American Medical Association}
}

@inproceedings{bevan2022detecting,
  title        = {Detecting melanoma fairly: Skin tone detection and debiasing for skin lesion classification},
  author       = {Bevan, Peter J and Atapour-Abarghouei, Amir},
  booktitle    = {MICCAI Workshop on Domain Adaptation and Representation Transfer},
  pages        = {1--11},
  year         = {2022},
  organization = {Springer}
}

@inproceedings{ansari2024algorithmic,
  title        = {Algorithmic Fairness in Lesion Classification by Mitigating Class Imbalance and Skin Tone Bias},
  author       = {Ansari, Faizanuddin and Chakraborti, Tapabrata and Das, Swagatam},
  booktitle    = {International Conference on Medical Image Computing and Computer-Assisted Intervention},
  pages        = {373--382},
  year         = {2024},
  organization = {Springer}
}

@inproceedings{bie2024xcoop,
  title        = {Xcoop: Explainable prompt learning for computer-aided diagnosis via concept-guided context optimization},
  author       = {Bie, Yequan and Luo, Luyang and Chen, Zhixuan and Chen, Hao},
  booktitle    = {International Conference on Medical Image Computing and Computer-Assisted Intervention},
  pages        = {773--783},
  year         = {2024},
  organization = {Springer}
}

@inproceedings{pahde2023reveal,
  title        = {Reveal to revise: An explainable ai life cycle for iterative bias correction of deep models},
  author       = {Pahde, Frederik and Dreyer, Maximilian and Samek, Wojciech and Lapuschkin, Sebastian},
  booktitle    = {International Conference on Medical Image Computing and Computer-Assisted Intervention},
  pages        = {596--606},
  year         = {2023},
  organization = {Springer}
}

@inproceedings{wang2024majority,
  title        = {From Majority to Minority: A Diffusion-based Augmentation for Underrepresented Groups in Skin Lesion Analysis},
  author       = {Wang, Janet and Chung, Yunsung and Ding, Zhengming and Hamm, Jihun},
  booktitle    = {International Conference on Medical Image Computing and Computer-Assisted Intervention},
  pages        = {14--23},
  year         = {2024},
  organization = {Springer}
}

@inproceedings{carrasco2022assessing,
  title        = {Assessing gan-based generative modeling on skin lesions images},
  author       = {Carrasco Limeros, Sandra and Majchrowska, Sylwia and Zoubi, Mohamad Khir and Ros{\'e}n, Anna and Suvilehto, Juulia and Sj{\"o}blom, Lisa and Kjellberg, Magnus},
  booktitle    = {Machine Intelligence and Digital Interaction Conference},
  pages        = {93--102},
  year         = {2022},
  organization = {Springer Nature Switzerland Cham}
}

@article{tai2024cancer,
  title   = {Cancer-Net SCa-Synth: An Open Access Synthetically Generated 2D Skin Lesion Dataset for Skin Cancer Classification},
  author  = {Tai, Chi-en Amy and Ding, Oustan and Wong, Alexander},
  journal = {arXiv preprint arXiv:2411.05269},
  year    = {2024}
}

@article{baur2018melanogans,
  title   = {MelanoGANs: High Resolution Skin Lesion Synthesis with GANs},
  author  = {Baur, Christoph and Albarqouni, Shadi and Navab, Nassir},
  journal = {arXiv preprint arXiv:1804.04338},
  year    = {2018}
}

@article{radford2015unsupervised,
  title   = {Unsupervised representation learning with deep convolutional generative adversarial networks},
  author  = {Radford, Alec and Metz, Luke and Chintala, Soumith},
  journal = {arXiv preprint arXiv:1511.06434},
  year    = {2015}
}

@article{denton2015deep,
  title   = {Deep generative image models using a laplacian pyramid of adversarial networks},
  author  = {Denton, Emily L and Chintala, Soumith and Fergus, Rob and others},
  journal = {Advances in neural information processing systems},
  volume  = {28},
  year    = {2015}
}

@article{ghorbani2019dermgan,
  title   = {DermGAN: Synthetic Generation of Clinical Skin Images with Pathology},
  author  = {Ghorbani, Amirata and Natarajan, Vivek and Coz, David and Liu, Yuan},
  journal = {arXiv preprint arXiv:1911.08716},
  year    = {2019}
}

@inproceedings{karras2019style,
  title     = {A style-based generator architecture for generative adversarial networks},
  author    = {Karras, Tero and Laine, Samuli and Aila, Timo},
  booktitle = {Proceedings of the IEEE/CVF conference on computer vision and pattern recognition},
  pages     = {4401--4410},
  year      = {2019}
}

@article{karras2020training,
  title   = {Training generative adversarial networks with limited data},
  author  = {Karras, Tero and Aittala, Miika and Hellsten, Janne and Laine, Samuli and Lehtinen, Jaakko and Aila, Timo},
  journal = {Advances in neural information processing systems},
  volume  = {33},
  pages   = {12104--12114},
  year    = {2020}
}

@inproceedings{rombach2022high,
  title     = {High-resolution image synthesis with latent diffusion models},
  author    = {Rombach, Robin and Blattmann, Andreas and Lorenz, Dominik and Esser, Patrick and Ommer, Bj{\"o}rn},
  booktitle = {Proceedings of the IEEE/CVF conference on computer vision and pattern recognition},
  pages     = {10684--10695},
  year      = {2022}
}

@article{bissoto2019skin,
  title   = {Skin Lesion Synthesis with Generative Adversarial Networks},
  author  = {Bissoto, Alceu and Perez, Fábio and Valle, Eduardo and Avila, Sandra},
  journal = {arXiv preprint arXiv:1902.03253},
  year    = {2019}
}

@article{sachdeva2009fitzpatrick,
  title     = {Fitzpatrick skin typing: Applications in dermatology},
  author    = {Sachdeva, Silonie},
  journal   = {Indian journal of dermatology, venereology and leprology},
  volume    = {75},
  pages     = {93},
  year      = {2009},
  publisher = {scientific scholar}
}

@article{ma2023effnet,
  title     = {EFFNet: A skin cancer classification model based on feature fusion and random forests},
  author    = {Ma, Xiaopu and Shan, Jiangdan and Ning, Fei and Li, Wentao and Li, He},
  journal   = {Plos one},
  volume    = {18},
  number    = {10},
  pages     = {e0293266},
  year      = {2023},
  publisher = {Public Library of Science San Francisco, CA USA}
}

@inproceedings{ruiz2023dreambooth,
  title     = {Dreambooth: Fine tuning text-to-image diffusion models for subject-driven generation},
  author    = {Ruiz, Nataniel and Li, Yuanzhen and Jampani, Varun and Pritch, Yael and Rubinstein, Michael and Aberman, Kfir},
  booktitle = {Proceedings of the IEEE/CVF conference on computer vision and pattern recognition},
  pages     = {22500--22510},
  year      = {2023}
}

@inproceedings{radford2021learning,
  title     = {Learning Transferable Visual Models From Natural Language Supervision},
  author    = {Radford, Alec and Kim, Jong Wook and Hallacy, Luke and Ramesh, Aditya and Goh, Gabriel and Agarwal, Sandhini and Sastry, Girish and Askell, Amanda and Mishkin, Pamela and Clark, Jack and Krueger, Gretchen and Sutskever, Ilya},
  booktitle = {Proceedings of the 38th International Conference on Machine Learning (ICML)},
  year      = {2021}
}

@article{loshchilov2019decoupled,
  title   = {Decoupled Weight Decay Regularization},
  author  = {Loshchilov, Ilya and Hutter, Frank},
  journal = {International Conference on Learning Representations (ICLR)},
  year    = {2019},
  url     = {https://arxiv.org/abs/1711.05101}
}

@article{oquab2023dinov2,
  title   = {Dinov2: Learning robust visual features without supervision},
  author  = {Oquab, Maxime and Darcet, Timoth{\'e}e and Moutakanni, Th{\'e}o and Vo, Huy and Szafraniec, Marc and Khalidov, Vasil and Fernandez, Pierre and Haziza, Daniel and Massa, Francisco and El-Nouby, Alaaeldin and others},
  journal = {arXiv preprint arXiv:2304.07193},
  year    = {2023}
}

@inproceedings{he2022masked,
  title     = {Masked autoencoders are scalable vision learners},
  author    = {He, Kaiming and Chen, Xinlei and Xie, Saining and Li, Yanghao and Doll{\'a}r, Piotr and Girshick, Ross},
  booktitle = {Proceedings of the IEEE/CVF conference on computer vision and pattern recognition},
  pages     = {16000--16009},
  year      = {2022}
}

@inproceedings{yao2025reconstruction,
  title     = {Reconstruction vs. generation: Taming optimization dilemma in latent diffusion models},
  author    = {Yao, Jingfeng and Yang, Bin and Wang, Xinggang},
  booktitle = {Proceedings of the Computer Vision and Pattern Recognition Conference},
  pages     = {15703--15712},
  year      = {2025}
}

@article{gutman2016skin,
  title   = {Skin lesion analysis toward melanoma detection: A challenge at the international symposium on biomedical imaging (ISBI) 2016, hosted by the international skin imaging collaboration (ISIC)},
  author  = {Gutman, David and Codella, Noel CF and Celebi, Emre and Helba, Brian and Marchetti, Michael and Mishra, Nabin and Halpern, Allan},
  journal = {arXiv preprint arXiv:1605.01397},
  year    = {2016}
}

@inproceedings{codella2018skin,
  title        = {Skin lesion analysis toward melanoma detection: A challenge at the 2017 international symposium on biomedical imaging (isbi), hosted by the international skin imaging collaboration (isic)},
  author       = {Codella, Noel CF and Gutman, David and Celebi, M Emre and Helba, Brian and Marchetti, Michael A and Dusza, Stephen W and Kalloo, Aadi and Liopyris, Konstantinos and Mishra, Nabin and Kittler, Harald and others},
  booktitle    = {2018 IEEE 15th international symposium on biomedical imaging (ISBI 2018)},
  pages        = {168--172},
  year         = {2018},
  organization = {IEEE}
}

@article{codella2019skin,
  title   = {Skin lesion analysis toward melanoma detection 2018: A challenge hosted by the international skin imaging collaboration (isic)},
  author  = {Codella, Noel and Rotemberg, Veronica and Tschandl, Philipp and Celebi, M Emre and Dusza, Stephen and Gutman, David and Helba, Brian and Kalloo, Aadi and Liopyris, Konstantinos and Marchetti, Michael and others},
  journal = {arXiv preprint arXiv:1902.03368},
  year    = {2019}
}

@article{tschandl2018ham10000,
  title     = {The HAM10000 dataset, a large collection of multi-source dermatoscopic images of common pigmented skin lesions},
  author    = {Tschandl, Philipp and Rosendahl, Cliff and Kittler, Harald},
  journal   = {Scientific data},
  volume    = {5},
  number    = {1},
  pages     = {1--9},
  year      = {2018},
  publisher = {Nature Publishing Group}
}

@article{hernandez2024bcn20000,
  title     = {Bcn20000: Dermoscopic lesions in the wild},
  author    = {Hern{\'a}ndez-P{\'e}rez, Carlos and Combalia, Marc and Podlipnik, Sebastian and Codella, Noel CF and Rotemberg, Veronica and Halpern, Allan C and Reiter, Ofer and Carrera, Cristina and Barreiro, Alicia and Helba, Brian and others},
  journal   = {Scientific data},
  volume    = {11},
  number    = {1},
  pages     = {641},
  year      = {2024},
  publisher = {Nature Publishing Group UK London}
}

@article{rotemberg2021patient,
  title     = {A patient-centric dataset of images and metadata for identifying melanomas using clinical context},
  author    = {Rotemberg, Veronica and Kurtansky, Nicholas and Betz-Stablein, Brigid and Caffery, Liam and Chousakos, Emmanouil and Codella, Noel and Combalia, Marc and Dusza, Stephen and Guitera, Pascale and Gutman, David and others},
  journal   = {Scientific data},
  volume    = {8},
  number    = {1},
  pages     = {34},
  year      = {2021},
  publisher = {Nature Publishing Group UK London}
}

@article{kawahara2018seven,
  title     = {Seven-point checklist and skin lesion classification using multitask multimodal neural nets},
  author    = {Kawahara, Jeremy and Daneshvar, Sara and Argenziano, Giuseppe and Hamarneh, Ghassan},
  journal   = {IEEE journal of biomedical and health informatics},
  volume    = {23},
  number    = {2},
  pages     = {538--546},
  year      = {2018},
  publisher = {IEEE}
}

@inproceedings{mallya2022deep,
  title        = {Deep multimodal guidance for medical image classification},
  author       = {Mallya, Mayur and Hamarneh, Ghassan},
  booktitle    = {International Conference on Medical Image Computing and Computer-Assisted Intervention},
  pages        = {298--308},
  year         = {2022},
  organization = {Springer}
}

@article{zunair2020melanoma,
  title     = {Melanoma detection using adversarial training and deep transfer learning},
  author    = {Zunair, Hasib and Hamza, A Ben},
  journal   = {Physics in Medicine \& Biology},
  year      = {2020},
  publisher = {IOP Publishing}
}

@article{gessert2020skin,
  author   = {Gessert, Nils and Sentker, Thilo and Madesta, Frederic and Schmitz, Rüdiger and Kniep, Helge and Baltruschat, Ivo and Werner, René and Schlaefer, Alexander},
  journal  = {IEEE Transactions on Biomedical Engineering},
  title    = {Skin Lesion Classification Using CNNs With Patch-Based Attention and Diagnosis-Guided Loss Weighting},
  year     = {2020},
  volume   = {67},
  number   = {2},
  pages    = {495-503},
  doi      = {10.1109/TBME.2019.2915839}
}

@article{panda2021comparative,
  title   = {Comparative study on different Deep Learning models for Skin Lesion Classification using transfer learning approach},
  author  = {Panda, Saswat and Tiwari, Abhishek Sunil and Prusty, Manas Ranjan},
  journal = {International Journal of Scientific and Research Publications},
  volume  = {11},
  number  = {1},
  pages   = {219--32},
  year    = {2021}
}

@inproceedings{sheng2024data,
  title        = {Data-Algorithm-Architecture Co-Optimization for Fair Neural Networks on Skin Lesion Dataset},
  author       = {Sheng, Yi and Yang, Junhuan and Li, Jinyang and Alaina, James and Xu, Xiaowei and Shi, Yiyu and Hu, Jingtong and Jiang, Weiwen and Yang, Lei},
  booktitle    = {International Conference on Medical Image Computing and Computer-Assisted Intervention},
  pages        = {153--163},
  year         = {2024},
  organization = {Springer}
}

@inproceedings{aayushman2024fair,
  title        = {Fair and Accurate Skin Disease Image Classification by Alignment with Clinical Labels},
  author       = {Aayushman and Gaddey, Hemanth and Mittal, Vidhi and Chawla, Manisha and Gupta, Gagan Raj},
  booktitle    = {International Conference on Medical Image Computing and Computer-Assisted Intervention},
  pages        = {394--404},
  year         = {2024},
  organization = {Springer}
}

@article{sabir2024classification,
  title     = {Classification of melanoma skin Cancer based on Image Data Set using different neural networks},
  author    = {Sabir, Rukhsar and Mehmood, Tahir},
  journal   = {Scientific Reports},
  volume    = {14},
  number    = {1},
  pages     = {29704},
  year      = {2024},
  publisher = {Nature Publishing Group UK London}
}

@article{chao2017smartphone,
  title   = {Smartphone-based applications for skin monitoring and melanoma detection.},
  author  = {Chao, Elizabeth and Meenan, Chelsea K and Ferris, Laura K},
  journal = {Dermatologic clinics},
  volume  = {35},
  number  = {4},
  pages   = {551--557},
  year    = {2017}
}

@article{rat2018use,
  title     = {Use of smartphones for early detection of melanoma: systematic review},
  author    = {Rat, C{\'e}dric and Hild, Sandrine and Rault S{\'e}randour, Julie and Gaultier, Aur{\'e}lie and Quereux, Gaelle and Dreno, Brigitte and Nguyen, Jean-Michel},
  journal   = {Journal of medical Internet research},
  volume    = {20},
  number    = {4},
  pages     = {e135},
  year      = {2018},
  publisher = {JMIR Publications Toronto, Canada}
}

@misc{european2019ethics,
  author       = {{European Union High-Level Expert Group on AI}},
  title        = {Ethics Guidelines for Trustworthy AI},
  organization = {European Commission},
  url          = {https://digital-strategy.ec.europa.eu/en/library/ethics-guidelines-trustworthy-ai},
  urldate      = {2024-08-07},
  year         = {2019}
}

@article{zhang2024vision,
  title     = {Vision-language models for vision tasks: A survey},
  author    = {Zhang, Jingyi and Huang, Jiaxing and Jin, Sheng and Lu, Shijian},
  journal   = {IEEE transactions on pattern analysis and machine intelligence},
  volume    = {46},
  number    = {8},
  pages     = {5625--5644},
  year      = {2024},
  publisher = {IEEE}
}

@inproceedings{hardt2016equality,
  author    = {Hardt, Moritz and Price, Eric and Srebro, Nathan},
  title     = {Equality of opportunity in supervised learning},
  year      = {2016},
  isbn      = {9781510838819},
  publisher = {Curran Associates Inc.},
  address   = {Red Hook, NY, USA},
  booktitle = {Proceedings of the 30th International Conference on Neural Information Processing Systems},
  pages     = {3323–3331},
  numpages  = {9},
  location  = {Barcelona, Spain},
  series    = {NIPS'16}
}

@InProceedings{pmlr-v136-kwon20a,
  title = 	 {Appropriate Evaluation of Diagnostic Utility of Machine Learning Algorithm Generated Images},
  author =       {Kwon, Young Joon and Toussie, Danielle and Azour, Lea and Concepcion, Jose and Eber, Corey and Reina, G. Anthony and Tang, Ping Tak Peter and Doshi, Amish H. and Oermann, Eric K. and Costa, Anthony B.},
  booktitle = 	 {Proceedings of the Machine Learning for Health NeurIPS Workshop},
  pages = 	 {179--193},
  year = 	 {2020},
  editor = 	 {Alsentzer, Emily and McDermott, Matthew B. A. and Falck, Fabian and Sarkar, Suproteem K. and Roy, Subhrajit and Hyland, Stephanie L.},
  volume = 	 {136},
  series = 	 {Proceedings of Machine Learning Research},
  month = 	 {11 Dec},
  publisher =    {PMLR},
  url = 	 {https://proceedings.mlr.press/v136/kwon20a.html}
}

@article{ho2020denoising,
  title={Denoising diffusion probabilistic models},
  author={Ho, Jonathan and Jain, Ajay and Abbeel, Pieter},
  journal={Advances in neural information processing systems},
  volume={33},
  pages={6840--6851},
  year={2020}
}

@inproceedings{peebles2023scalable,
  title={Scalable diffusion models with transformers},
  author={Peebles, William and Xie, Saining},
  booktitle={Proceedings of the IEEE/CVF international conference on computer vision},
  pages={4195--4205},
  year={2023}
}

@article{goodfellow2020generative,
  title={Generative adversarial networks},
  author={Goodfellow, Ian and Pouget-Abadie, Jean and Mirza, Mehdi and Xu, Bing and Warde-Farley, David and Ozair, Sherjil and Courville, Aaron and Bengio, Yoshua},
  journal={Communications of the ACM},
  volume={63},
  number={11},
  pages={139--144},
  year={2020},
  publisher={ACM New York, NY, USA}
}

@misc{MILK10k_2025,
  author       = {{MILK Study Team}},
  title        = {MILK10k},
  howpublished = {ISIC Archive},
  year         = {2025},
  doi          = {10.34970/648456},
  url          = {https://doi.org/10.34970/648456}
}

@inproceedings{stanley2022disproportionate,
  author    = {Stanley, Emma A. M. and Wilms, Matthias and Forkert, Nils D.},
  title     = {Disproportionate Subgroup Impacts and Other Challenges of Fairness in Artificial Intelligence for Medical Image Analysis},
  year      = {2022},
  isbn      = {978-3-031-23222-0},
  publisher = {Springer-Verlag},
  address   = {Berlin, Heidelberg},
  url       = {https://doi.org/10.1007/978-3-031-23223-7_2},
  doi       = {10.1007/978-3-031-23223-7_2},
  booktitle = {Ethical and Philosophical Issues in Medical Imaging, Multimodal Learning and Fusion Across Scales for Clinical Decision Support, and Topological Data Analysis for Biomedical Imaging: 1st International Workshop, EPIMI 2022, 12th International Workshop, ML-CDS 2022, 2nd International Workshop, TDA4BiomedicalImaging, Held in Conjunction with MICCAI 2022, Singapore, September 18–22, 2022, Proceedings},
  pages     = {14–25},
  numpages  = {12},
  location  = {Singapore, Singapore}
}

@article{li2021estimating,
  title   = {Estimating and improving fairness with adversarial learning},
  author  = {Li, Xiaoxiao and Cui, Ziteng and Wu, Yifan and Gu, Lin and Harada, Tatsuya},
  journal = {arXiv preprint arXiv:2103.04243},
  year    = {2021}
}

@inproceedings{kinyanjui2020fairness,
  author    = {Kinyanjui, Newton M. and Odonga, Timothy and Cintas, Celia and Codella, Noel C. F. and Panda, Rameswar and Sattigeri, Prasanna and Varshney, Kush R.},
  title     = {Fairness of Classifiers Across Skin Tones in Dermatology},
  year      = {2020},
  isbn      = {978-3-030-59724-5},
  publisher = {Springer-Verlag},
  address   = {Berlin, Heidelberg},
  url       = {https://doi.org/10.1007/978-3-030-59725-2_31},
  doi       = {10.1007/978-3-030-59725-2_31},
  booktitle = {Medical Image Computing and Computer Assisted Intervention – MICCAI 2020: 23rd International Conference, Lima, Peru, October 4–8, 2020, Proceedings, Part VI},
  pages     = {320–329},
  numpages  = {10},
  location  = {Lima, Peru}
}

@article{rolnick2017deep,
  title={Deep learning is robust to massive label noise},
  author={Rolnick, David and Veit, Andreas and Belongie, Serge and Shavit, Nir},
  journal={arXiv preprint arXiv:1705.10694},
  year={2017}
}

@inproceedings{nejadgholi2022towards,
  title={Towards procedural fairness: Uncovering biases in how a toxic language classifier uses sentiment information},
  author={Nejadgholi, Isar and Balkir, Esma and Fraser, Kathleen C and Kiritchenko, Svetlana},
  booktitle={Proceedings of the fifth BlackboxNLP workshop on analyzing and interpreting neural networks for NLP},
  pages={225--237},
  year={2022}
}

@article{pendyala2022concept,
  title={Concept-based explanations for tabular data},
  author={Pendyala, Varsha and Choi, Jihye},
  journal={arXiv preprint arXiv:2209.05690},
  year={2022}
}

@article{tong2020investigating,
  title={Investigating bias in image classification using model explanations},
  author={Tong, Schrasing and Kagal, Lalana},
  journal={arXiv preprint arXiv:2012.05463},
  year={2020}
}

@inproceedings{kalb2023revisiting,
  title={Revisiting skin tone fairness in dermatological lesion classification},
  author={Kalb, Thorsten and Kushibar, Kaisar and Cintas, Celia and Lekadir, Karim and Diaz, Oliver and Osuala, Richard},
  booktitle={Workshop on Clinical Image-Based Procedures},
  pages={246--255},
  year={2023},
  organization={Springer}
}

@article{yang2024textbook,
 title={A Textbook Remedy for Domain Shifts: Knowledge Priors for Medical Image Analysis}, 
 author={Yue Yang and Mona Gandhi and Yufei Wang and Yifan Wu and Michael S. Yao and Chris Callison-Burch and James C. Gee and Mark Yatskar},
 journal={arXiv preprint arXiv:2405.14839},
 year={2024}
}
}

\end{document}


\maketitle

\section*{Analysis of sampling parameters}
To better understand the performance of our generative model we sweep the classifier-free guidance (CFG) scale and number of sampling steps, see \autoref{tab:fid_cfg_steps_ageavg}.
We find that the default sampling parameters are not optimal.
Instead, a lower CFG and higher number of steps produce the best FID (lower is better).
Therefore, we show synthetic images using these optimal parameters in the main paper.
For comparison, we display synthetic images with default sampling parameters in \autoref{fig:synth_melanoma_samples}.

\section*{Impact of data availability on generation quality}
Next, we are interested in the relationship between data availability and final generation quality.
To that end, we measure the FID of synthetic images grouped by age against real data and compute the corresponding available training data for said age group in \autoref{tab:fid_age_means}.
We find that the generation quality correlates well with the data availability.
Specifically, the best FID is achieved for age group 50 and 60, both represented more than 10\% in the training data.

\section*{Visualization of real testing data}
In \autoref{fig:melanoma_samples}, we visualize real testing data from the MILK10k \cite{MILK10k_2025} dataset.
Firstly, we observe that many combinations are simply not available thus making fairness assessment challenging.
Secondly, as seen in both the real and synthetic images, they do not become progressively darker with higher Fitzpatrick skin types.
Prior work \cite{tong2020investigating,kalb2023revisiting} shows that this intuitive assumption fails because lighting, camera settings, and lesion visibility heavily affect perceived darkness, leading expert-, crowd-, and algorithm-based annotations to diverge.
Due to this variability, raw pixel intensity is an unreliable indicator of skin type, and higher categories do not consistently appear darker, highlighting the need for calibrated annotations rather than appearance alone.

\section*{Nearest neighbours visualization}
\begin{figure*}[t]
\centering
\setlength{\tabcolsep}{1pt}

\begin{tabular}{@{}*{6}{c}@{}}
\multicolumn{6}{c}{\small Age 10, Female} \\[4pt]
\synimgPairs{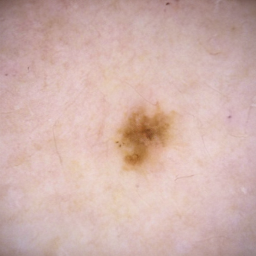} &
\synimgPairs{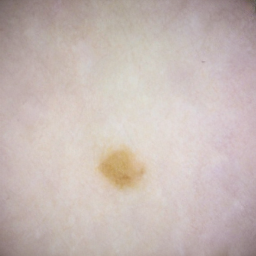} &
\synimgPairs{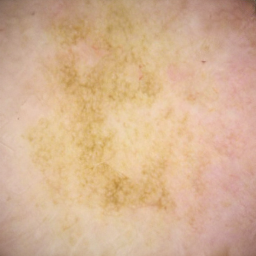} &
\synimgPairs{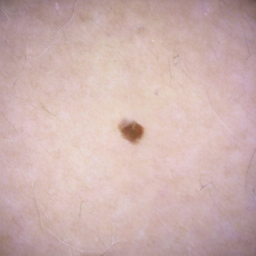} &
\synimgPairs{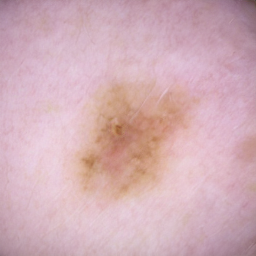} &
\synimgPairs{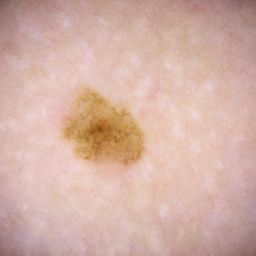} \\[2pt]
\realimgPairs{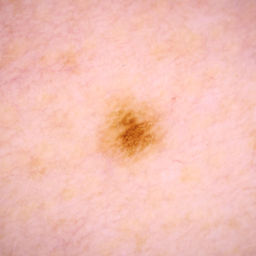} &
\realimgPairs{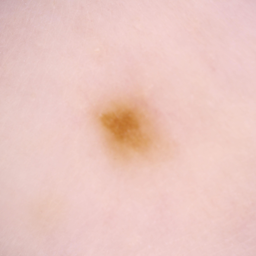} &
\realimgPairs{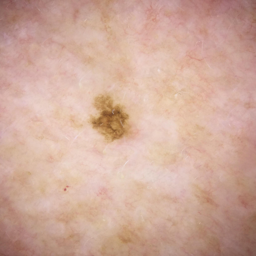} &
\realimgPairs{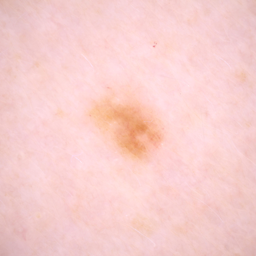} &
\realimgPairs{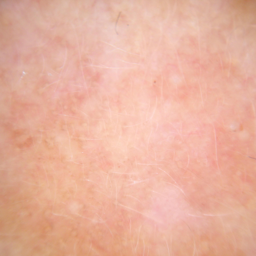} &
\realimgPairs{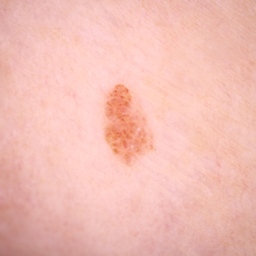} \\
\end{tabular}

\vspace{1em}

\begin{tabular}{@{}*{6}{c}@{}}
\multicolumn{6}{c}{\small Age 10, Male} \\[4pt]
\synimgPairs{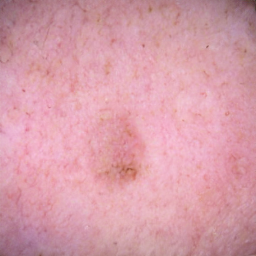} &
\synimgPairs{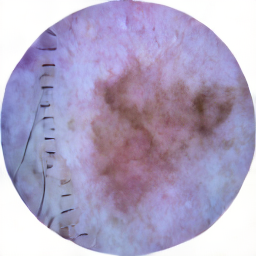} &
\synimgPairs{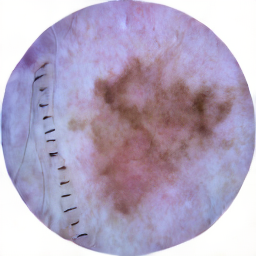} &
\synimgPairs{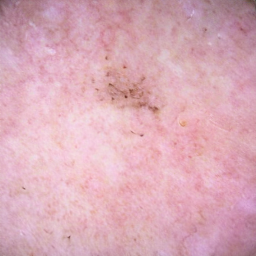} &
\synimgPairs{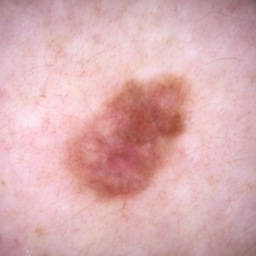} &
\synimgPairs{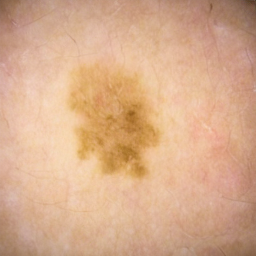} \\[2pt]
\realimgPairs{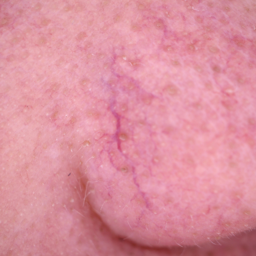} &
\realimgPairs{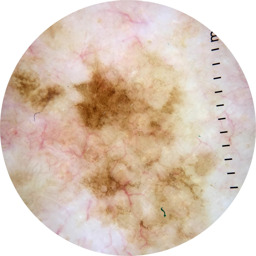} &
\realimgPairs{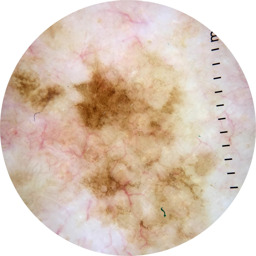} &
\realimgPairs{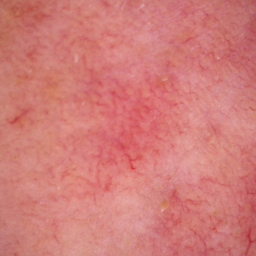} &
\realimgPairs{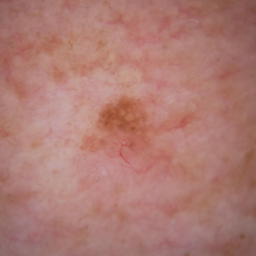} &
\realimgPairs{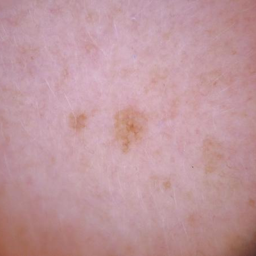} \\
\end{tabular}

\vspace{1em}

\begin{tabular}{@{}*{6}{c}@{}}
\multicolumn{6}{c}{\small Age 20, Female} \\[4pt]
\synimgPairs{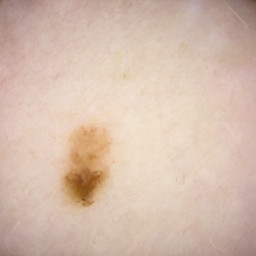} &
\synimgPairs{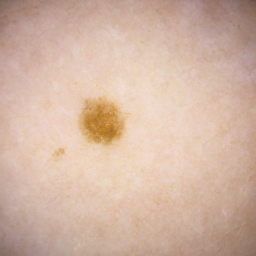} &
\synimgPairs{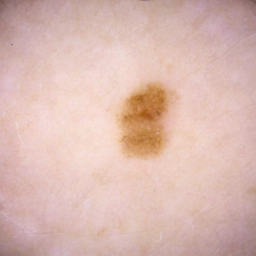} &
\synimgPairs{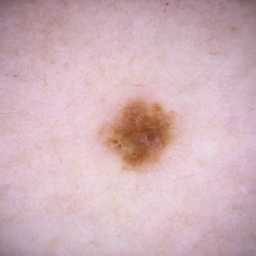} &
\synimgPairs{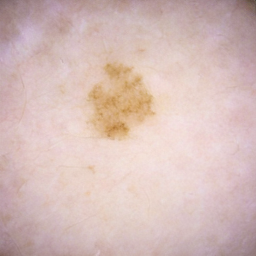} &
\synimgPairs{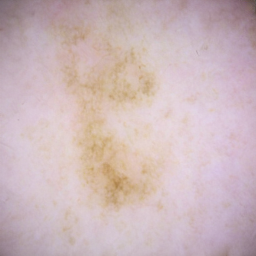} \\[2pt]
\realimgPairs{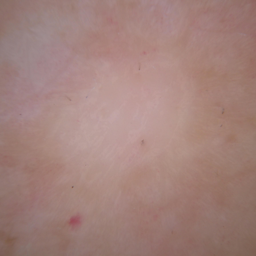} &
\realimgPairs{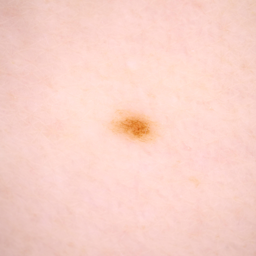} &
\realimgPairs{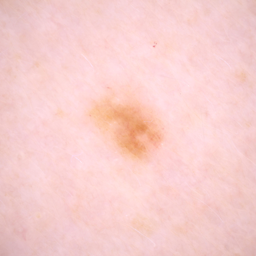} &
\realimgPairs{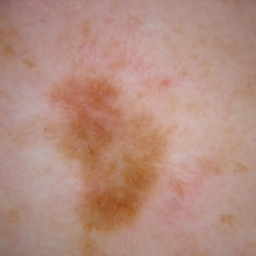} &
\realimgPairs{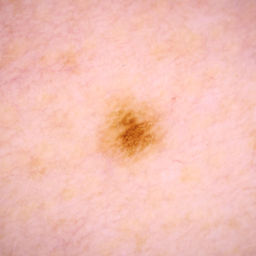} &
\realimgPairs{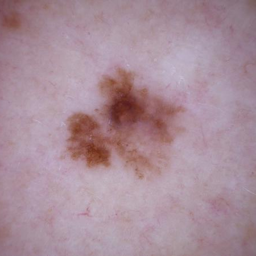} \\
\end{tabular}

\vspace{1em}

\begin{tabular}{@{}*{6}{c}@{}}
\multicolumn{6}{c}{\small Age 20, Male} \\[4pt]
\synimgPairs{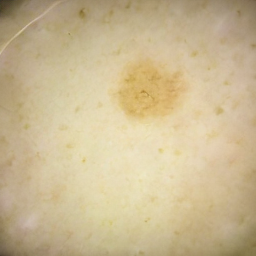} &
\synimgPairs{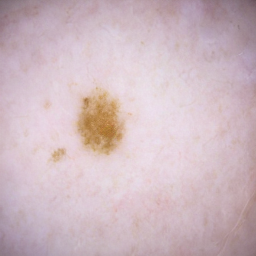} &
\synimgPairs{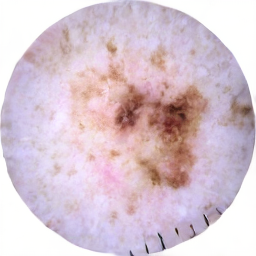} &
\synimgPairs{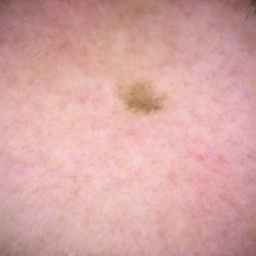} &
\synimgPairs{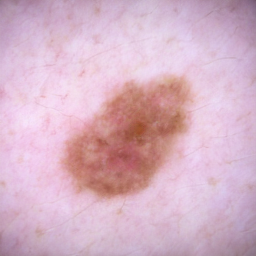} &
\synimgPairs{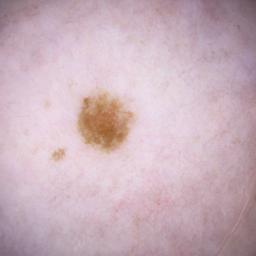} \\[2pt]
\realimgPairs{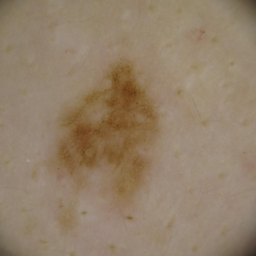} &
\realimgPairs{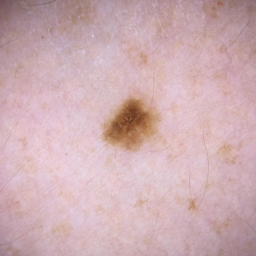} &
\realimgPairs{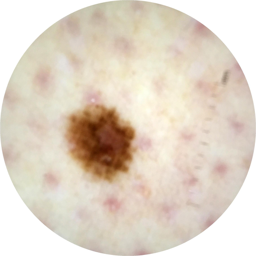} &
\realimgPairs{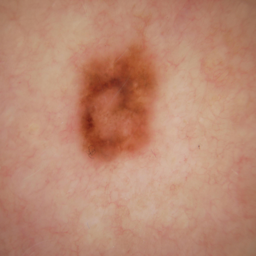} &
\realimgPairs{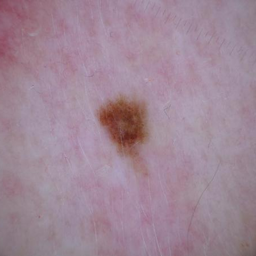} &
\realimgPairs{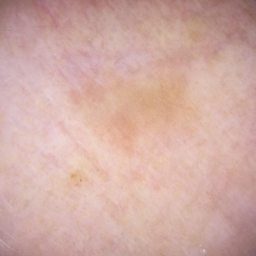} \\
\end{tabular}

\caption{Synthetic samples (top row in each block) and nearest real neighbours (bottom row) for ages 10 and 20. Columns represent skin type I to VI.}
\label{fig:synRealRows_10_20}
\end{figure*}

\begin{figure*}[t]
\centering
\setlength{\tabcolsep}{1pt}

\begin{tabular}{@{}*{6}{c}@{}}
\multicolumn{6}{c}{\small Age 30, Female} \\[4pt]
\synimgPairs{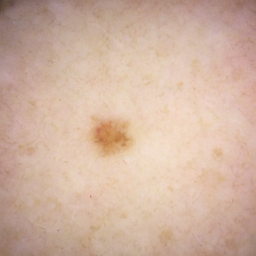} &
\synimgPairs{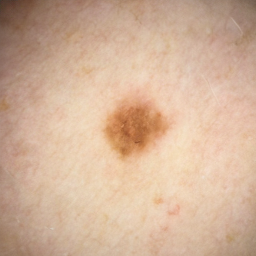} &
\synimgPairs{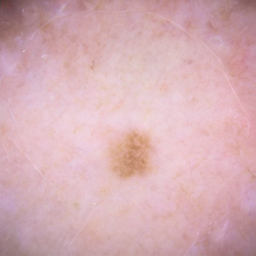} &
\synimgPairs{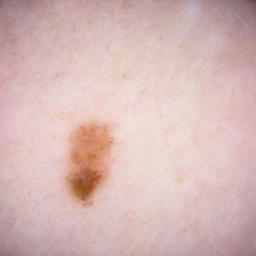} &
\synimgPairs{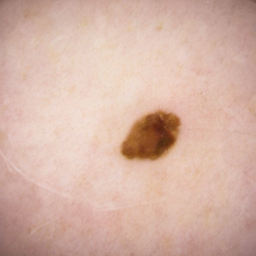} &
\synimgPairs{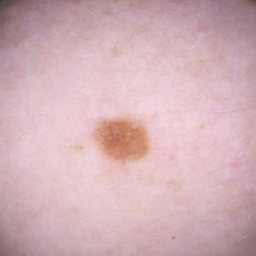} \\[2pt]
\realimgPairs{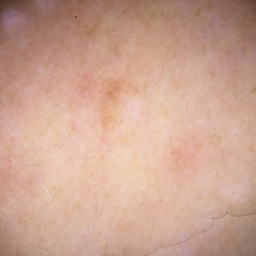} &
\realimgPairs{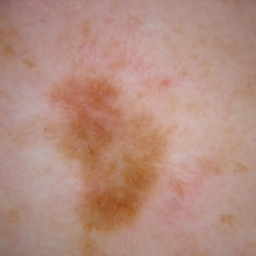} &
\realimgPairs{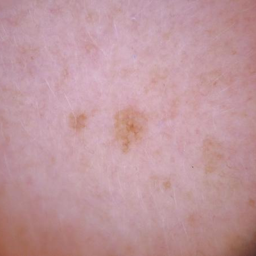} &
\realimgPairs{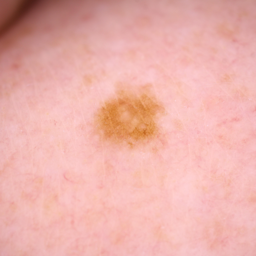} &
\realimgPairs{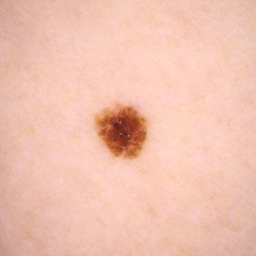} &
\realimgPairs{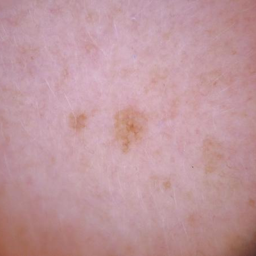} \\
\end{tabular}

\vspace{1em}

\begin{tabular}{@{}*{6}{c}@{}}
\multicolumn{6}{c}{\small Age 30, Male} \\[4pt]
\synimgPairs{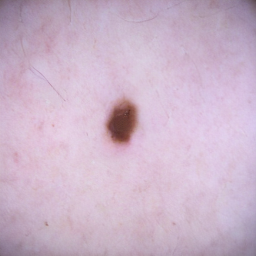} &
\synimgPairs{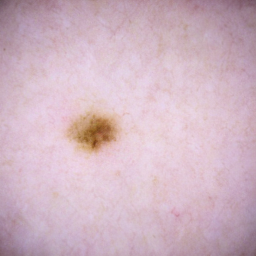} &
\synimgPairs{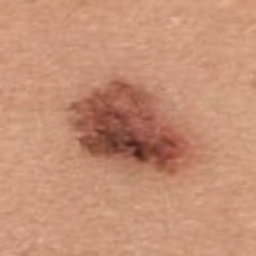} &
\synimgPairs{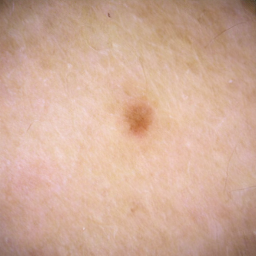} &
\synimgPairs{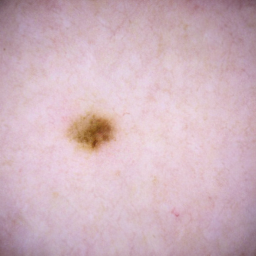} &
\synimgPairs{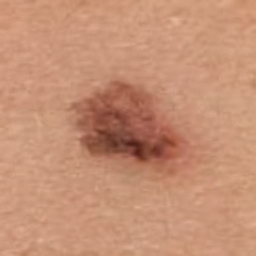} \\[2pt]
\realimgPairs{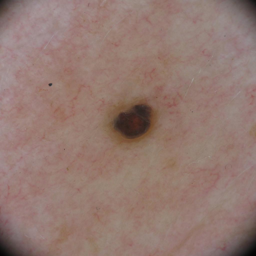} &
\realimgPairs{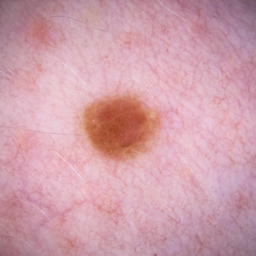} &
\realimgPairs{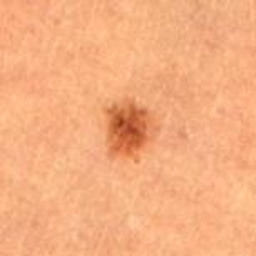} &
\realimgPairs{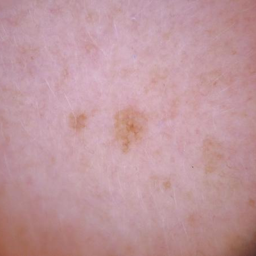} &
\realimgPairs{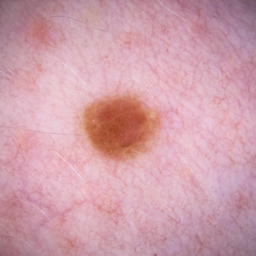} &
\realimgPairs{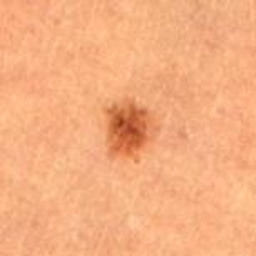} \\
\end{tabular}

\vspace{1em}

\begin{tabular}{@{}*{6}{c}@{}}
\multicolumn{6}{c}{\small Age 40, Female} \\[4pt]
\synimgPairs{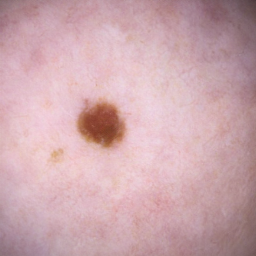} &
\synimgPairs{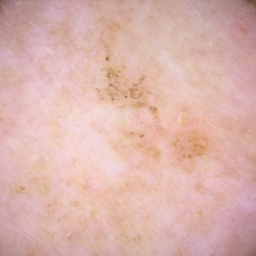} &
\synimgPairs{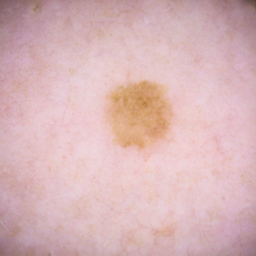} &
\synimgPairs{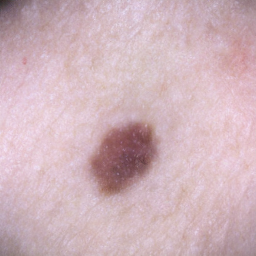} &
\synimgPairs{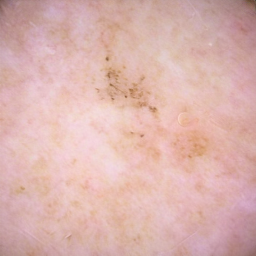} &
\synimgPairs{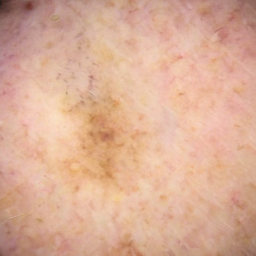} \\[2pt]
\realimgPairs{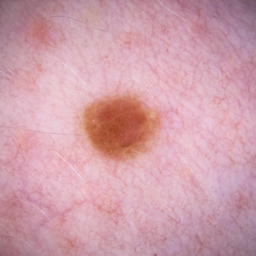} &
\realimgPairs{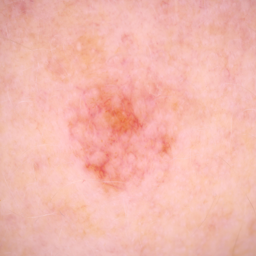} &
\realimgPairs{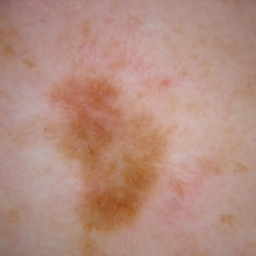} &
\realimgPairs{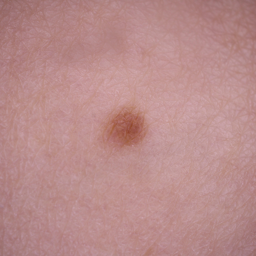} &
\realimgPairs{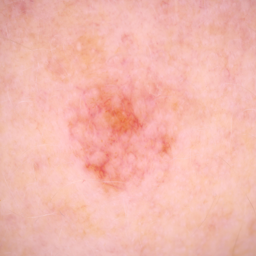} &
\realimgPairs{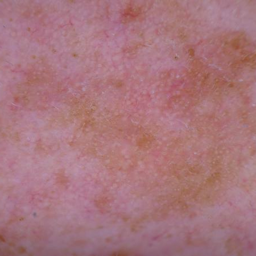} \\
\end{tabular}

\vspace{1em}

\begin{tabular}{@{}*{6}{c}@{}}
\multicolumn{6}{c}{\small Age 40, Male} \\[4pt]
\synimgPairs{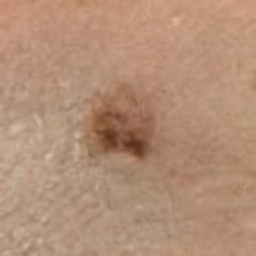} &
\synimgPairs{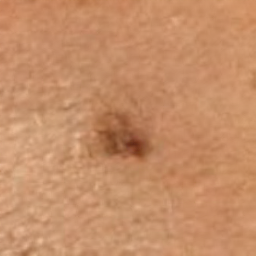} &
\synimgPairs{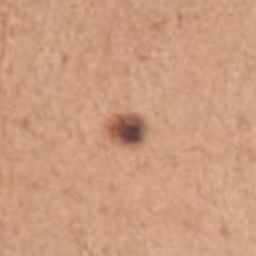} &
\synimgPairs{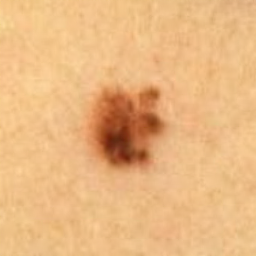} &
\synimgPairs{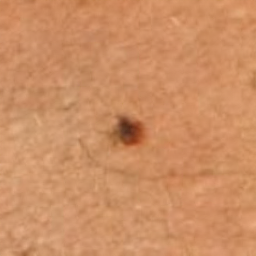} &
\synimgPairs{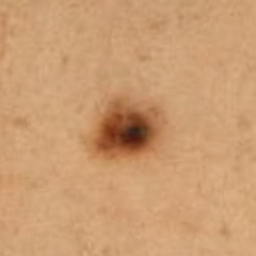} \\[2pt]
\realimgPairs{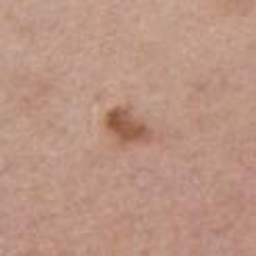} &
\realimgPairs{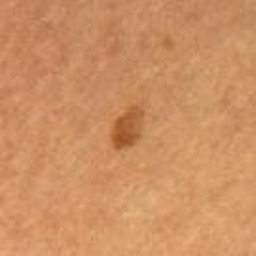} &
\realimgPairs{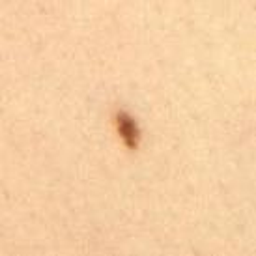} &
\realimgPairs{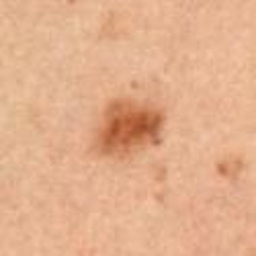} &
\realimgPairs{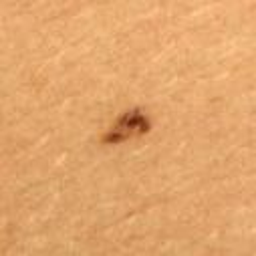} &
\realimgPairs{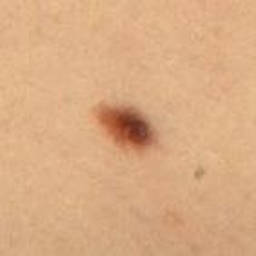} \\
\end{tabular}

\caption{Synthetic samples (top row in each block) and nearest real neighbours (bottom row) for ages 30 and 40. Columns represent skin type I to VI.}
\label{fig:synRealRows_30_40}
\end{figure*}

\begin{figure*}[t]
\centering
\setlength{\tabcolsep}{1pt}

\begin{tabular}{@{}*{6}{c}@{}}
\multicolumn{6}{c}{\small Age 50, Female} \\[4pt]
\synimgPairs{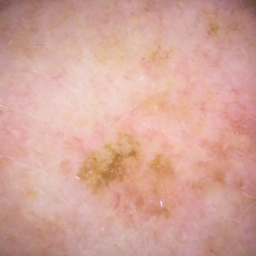} &
\synimgPairs{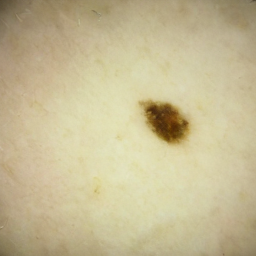} &
\synimgPairs{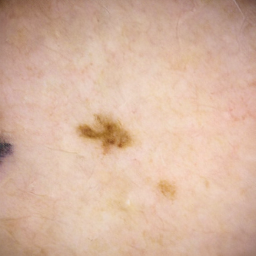} &
\synimgPairs{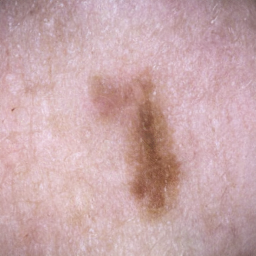} &
\synimgPairs{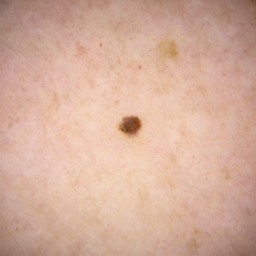} &
\synimgPairs{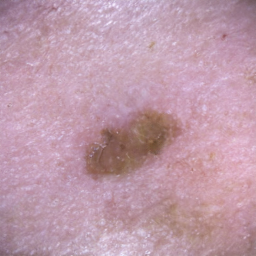} \\[2pt]
\realimgPairs{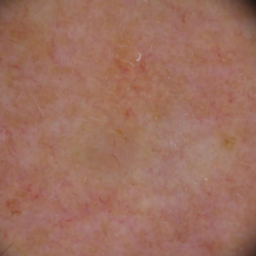} &
\realimgPairs{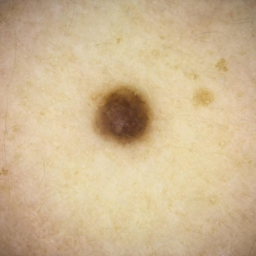} &
\realimgPairs{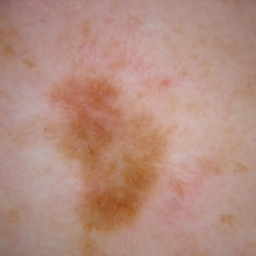} &
\realimgPairs{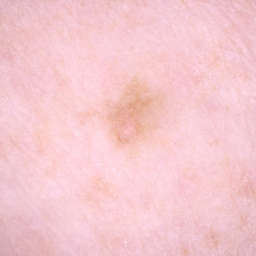} &
\realimgPairs{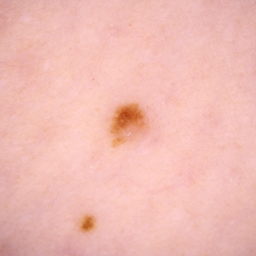} &
\realimgPairs{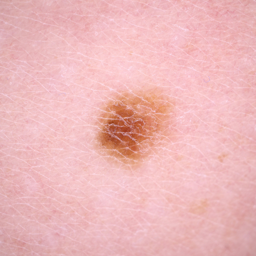} \\
\end{tabular}

\vspace{1em}

\begin{tabular}{@{}*{6}{c}@{}}
\multicolumn{6}{c}{\small Age 50, Male} \\[4pt]
\synimgPairs{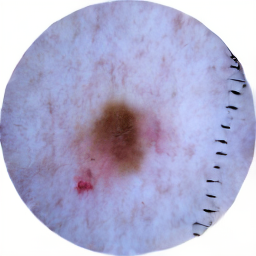} &
\synimgPairs{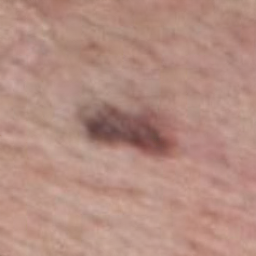} &
\synimgPairs{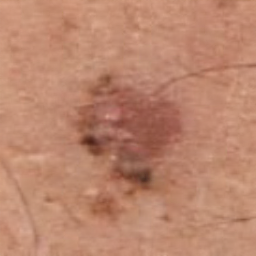} &
\synimgPairs{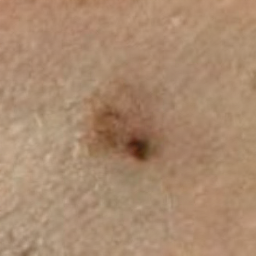} &
\synimgPairs{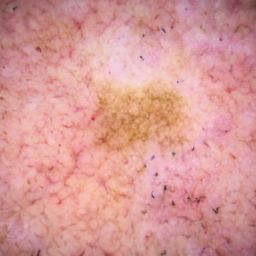} &
\synimgPairs{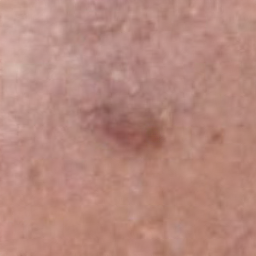} \\[2pt]
\realimgPairs{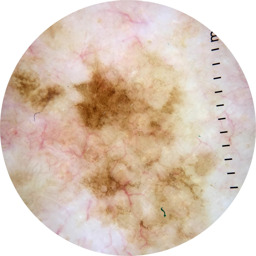} &
\realimgPairs{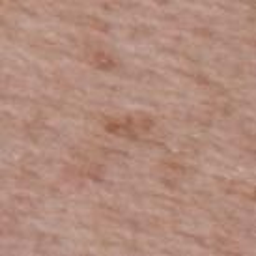} &
\realimgPairs{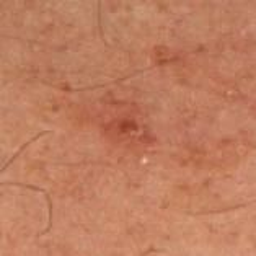} &
\realimgPairs{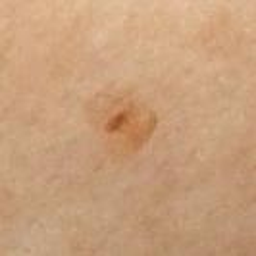} &
\realimgPairs{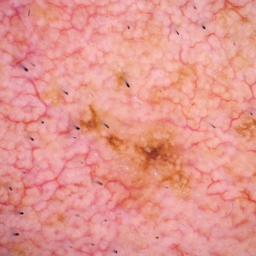} &
\realimgPairs{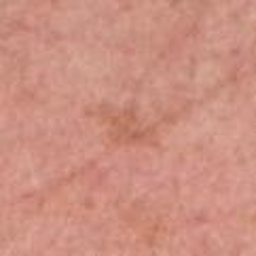} \\
\end{tabular}

\vspace{1em}

\begin{tabular}{@{}*{6}{c}@{}}
\multicolumn{6}{c}{\small Age 60, Female} \\[4pt]
\synimgPairs{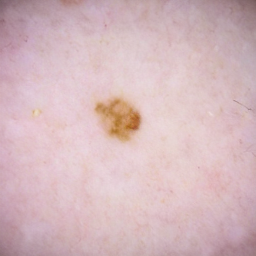} &
\synimgPairs{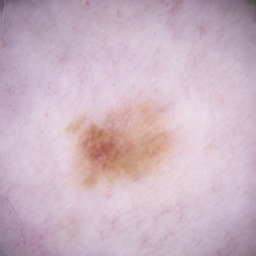} &
\synimgPairs{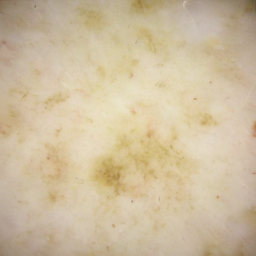} &
\synimgPairs{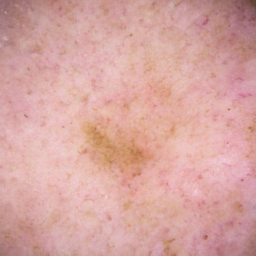} &
\synimgPairs{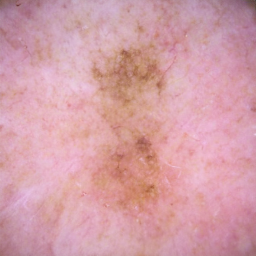} &
\synimgPairs{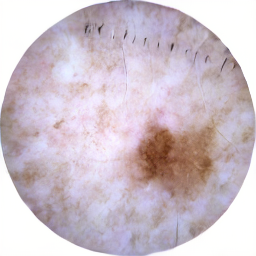} \\[2pt]
\realimgPairs{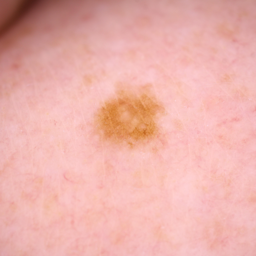} &
\realimgPairs{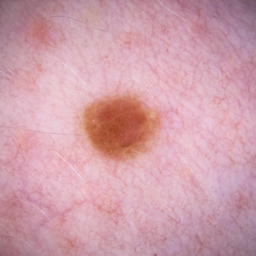} &
\realimgPairs{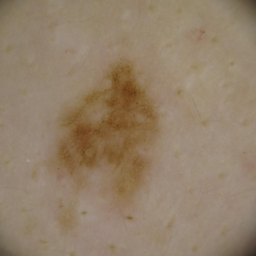} &
\realimgPairs{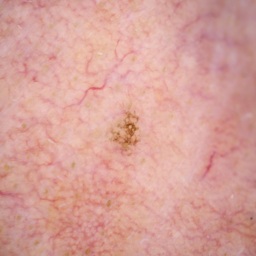} &
\realimgPairs{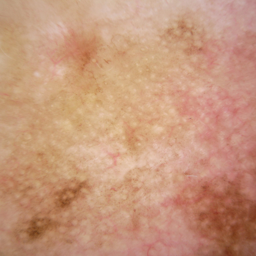} &
\realimgPairs{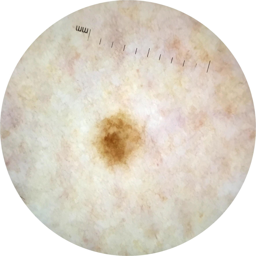} \\
\end{tabular}

\vspace{1em}

\begin{tabular}{@{}*{6}{c}@{}}
\multicolumn{6}{c}{\small Age 60, Male} \\[4pt]
\synimgPairs{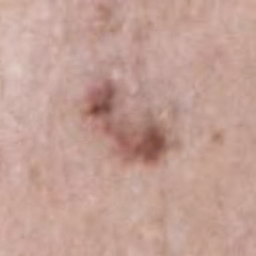} &
\synimgPairs{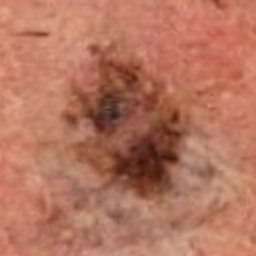} &
\synimgPairs{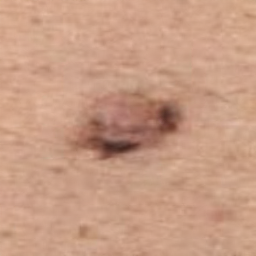} &
\synimgPairs{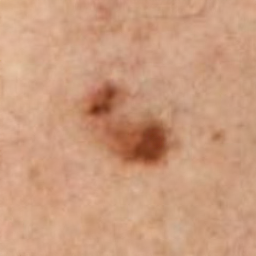} &
\synimgPairs{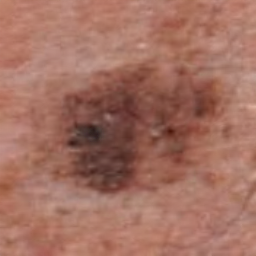} &
\synimgPairs{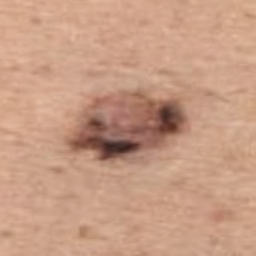} \\[2pt]
\realimgPairs{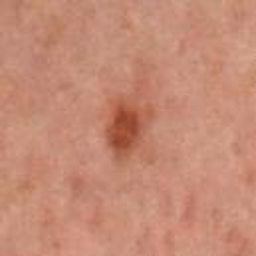} &
\realimgPairs{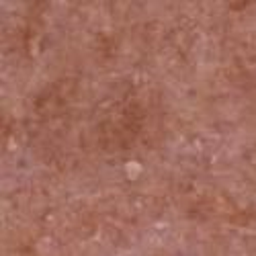} &
\realimgPairs{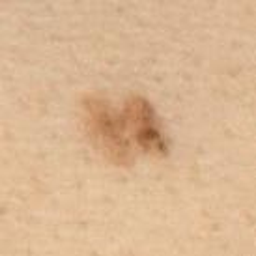} &
\realimgPairs{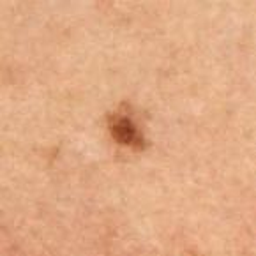} &
\realimgPairs{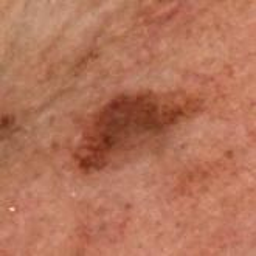} &
\realimgPairs{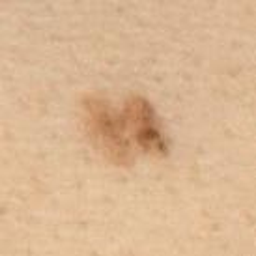} \\
\end{tabular}

\caption{Synthetic samples (top row in each block) and nearest real neighbours (bottom row) for ages 50 and 60. Columns represent skin type I to VI.}
\label{fig:synRealRows_50_60}
\end{figure*}

\begin{figure*}[t]
\centering
\setlength{\tabcolsep}{1pt}

\begin{tabular}{@{}*{6}{c}@{}}
\multicolumn{6}{c}{\small Age 70, Female} \\[4pt]
\synimgPairs{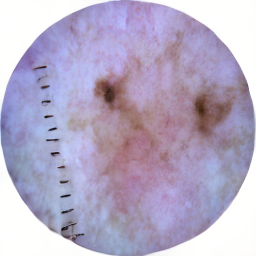} &
\synimgPairs{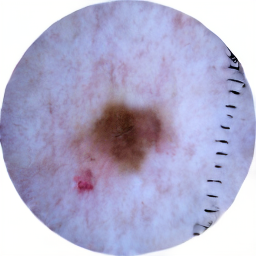} &
\synimgPairs{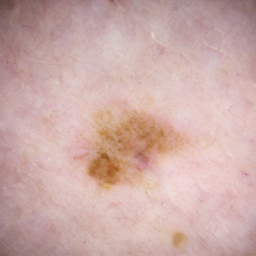} &
\synimgPairs{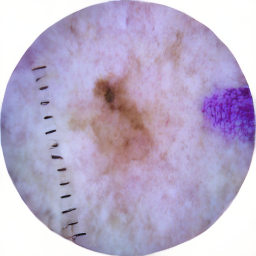} &
\synimgPairs{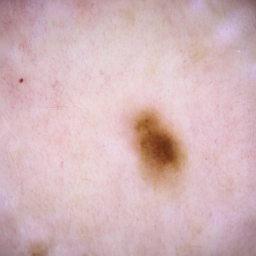} &
\synimgPairs{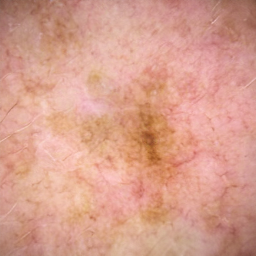} \\[2pt]
\realimgPairs{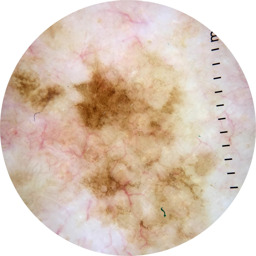} &
\realimgPairs{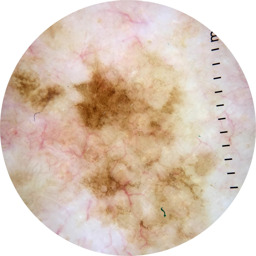} &
\realimgPairs{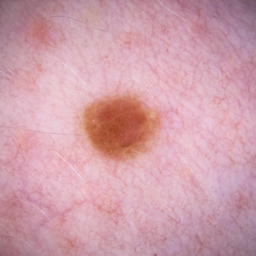} &
\realimgPairs{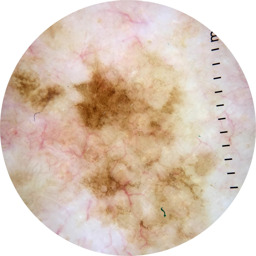} &
\realimgPairs{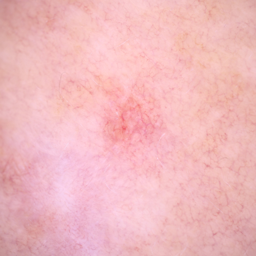} &
\realimgPairs{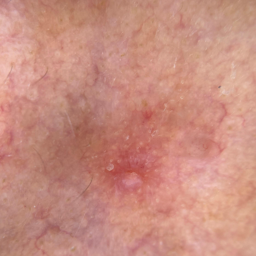} \\
\end{tabular}

\vspace{1em}

\begin{tabular}{@{}*{6}{c}@{}}
\multicolumn{6}{c}{\small Age 70, Male} \\[4pt]
\synimgPairs{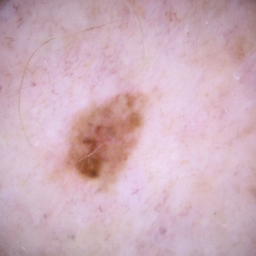} &
\synimgPairs{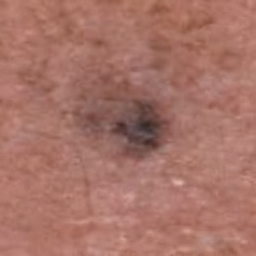} &
\synimgPairs{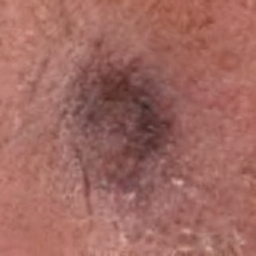} &
\synimgPairs{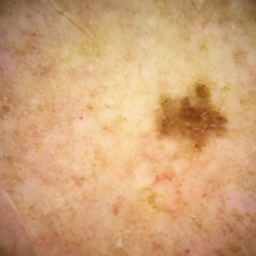} &
\synimgPairs{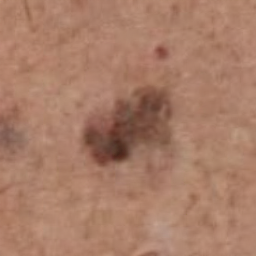} &
\synimgPairs{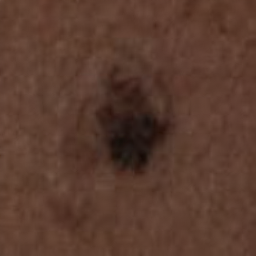} \\[2pt]
\realimgPairs{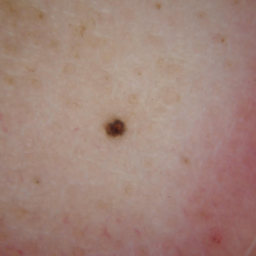} &
\realimgPairs{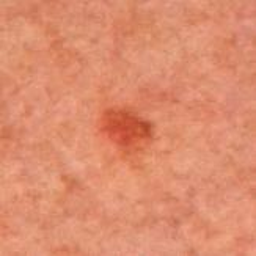} &
\realimgPairs{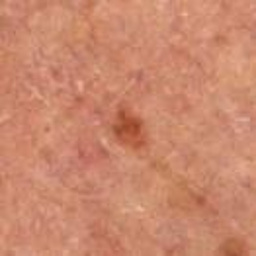} &
\realimgPairs{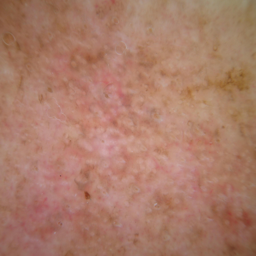} &
\realimgPairs{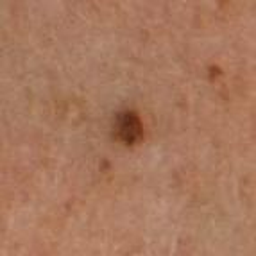} &
\realimgPairs{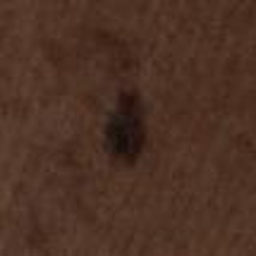} \\
\end{tabular}

\vspace{1em}

\begin{tabular}{@{}*{6}{c}@{}}
\multicolumn{6}{c}{\small Age 80, Female} \\[4pt]
\synimgPairs{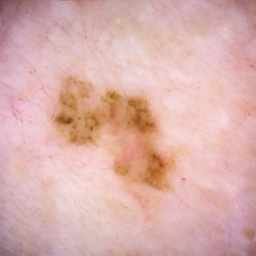} &
\synimgPairs{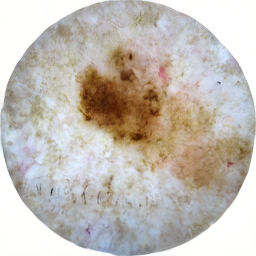} &
\synimgPairs{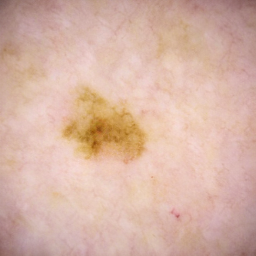} &
\synimgPairs{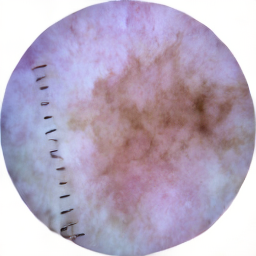} &
\synimgPairs{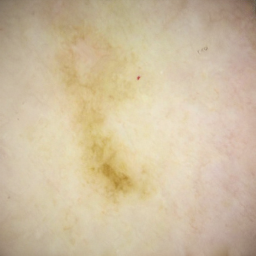} &
\synimgPairs{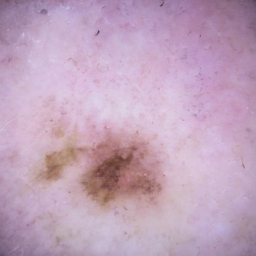} \\[2pt]
\realimgPairs{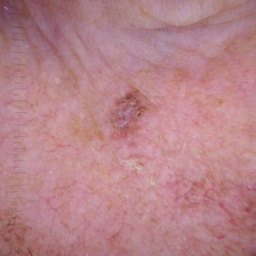} &
\realimgPairs{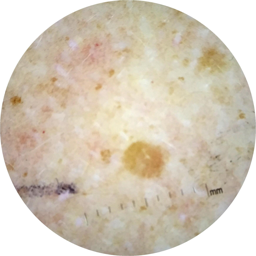} &
\realimgPairs{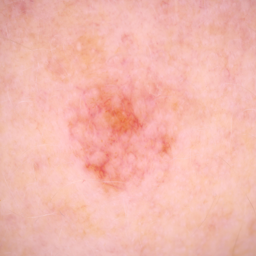} &
\realimgPairs{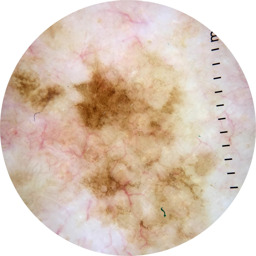} &
\realimgPairs{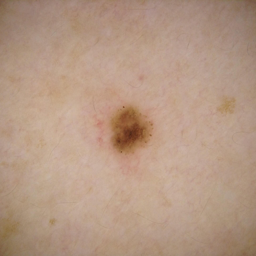} &
\realimgPairs{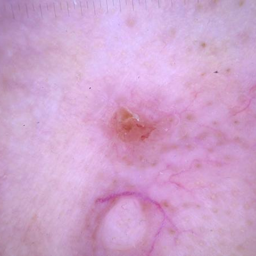} \\
\end{tabular}

\vspace{1em}

\begin{tabular}{@{}*{6}{c}@{}}
\multicolumn{6}{c}{\small Age 80, Male} \\[4pt]
\synimgPairs{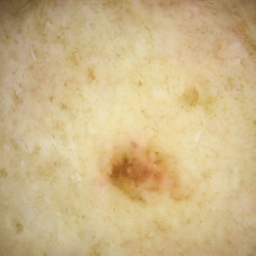} &
\synimgPairs{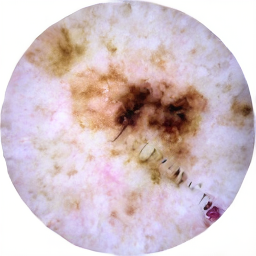} &
\synimgPairs{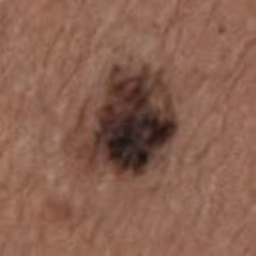} &
\synimgPairs{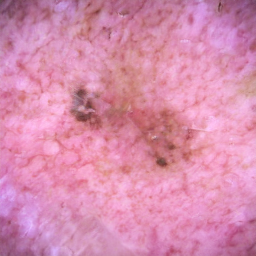} &
\synimgPairs{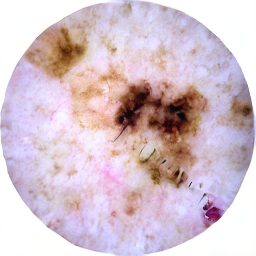} &
\synimgPairs{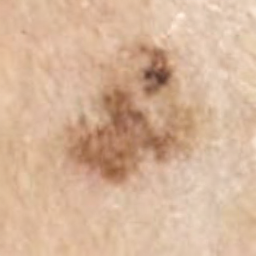} \\[2pt]
\realimgPairs{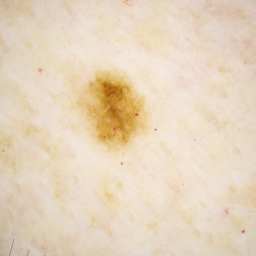} &
\realimgPairs{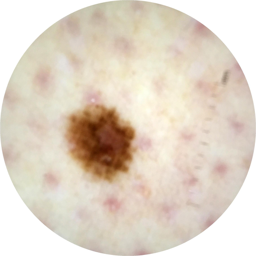} &
\realimgPairs{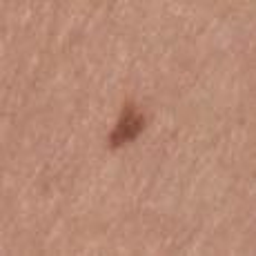} &
\realimgPairs{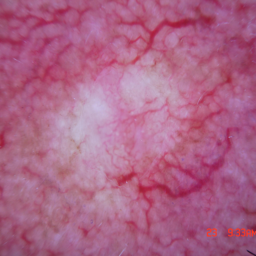} &
\realimgPairs{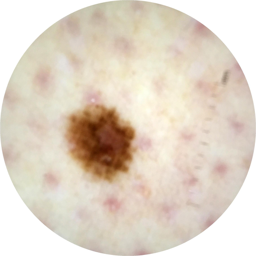} &
\realimgPairs{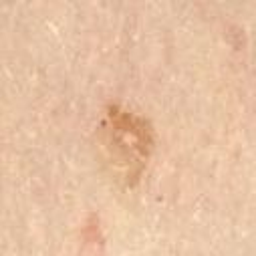} \\
\end{tabular}

\caption{Synthetic samples (top row in each block) and nearest real neighbours (bottom row) for ages 70 and 80. Columns represent skin type I to VI.}
\label{fig:synRealRows_70_80}
\end{figure*}

To further assess the capabilities of our generative model, we visualize synthetic images alongside their real nearest neighbours from the training set in \autoref{fig:synRealRows_10_20}–\autoref{fig:synRealRows_70_80}.
The generated samples are highly realistic yet clearly distinct from their nearest neighbours, indicating that the model does not merely memorize the training data.

\section*{Zero-shot classification comparison}
\begin{figure*}[t]
    \centering

    \begin{subfigure}{0.48\textwidth}
        \centering
        \includegraphics[width=\linewidth]{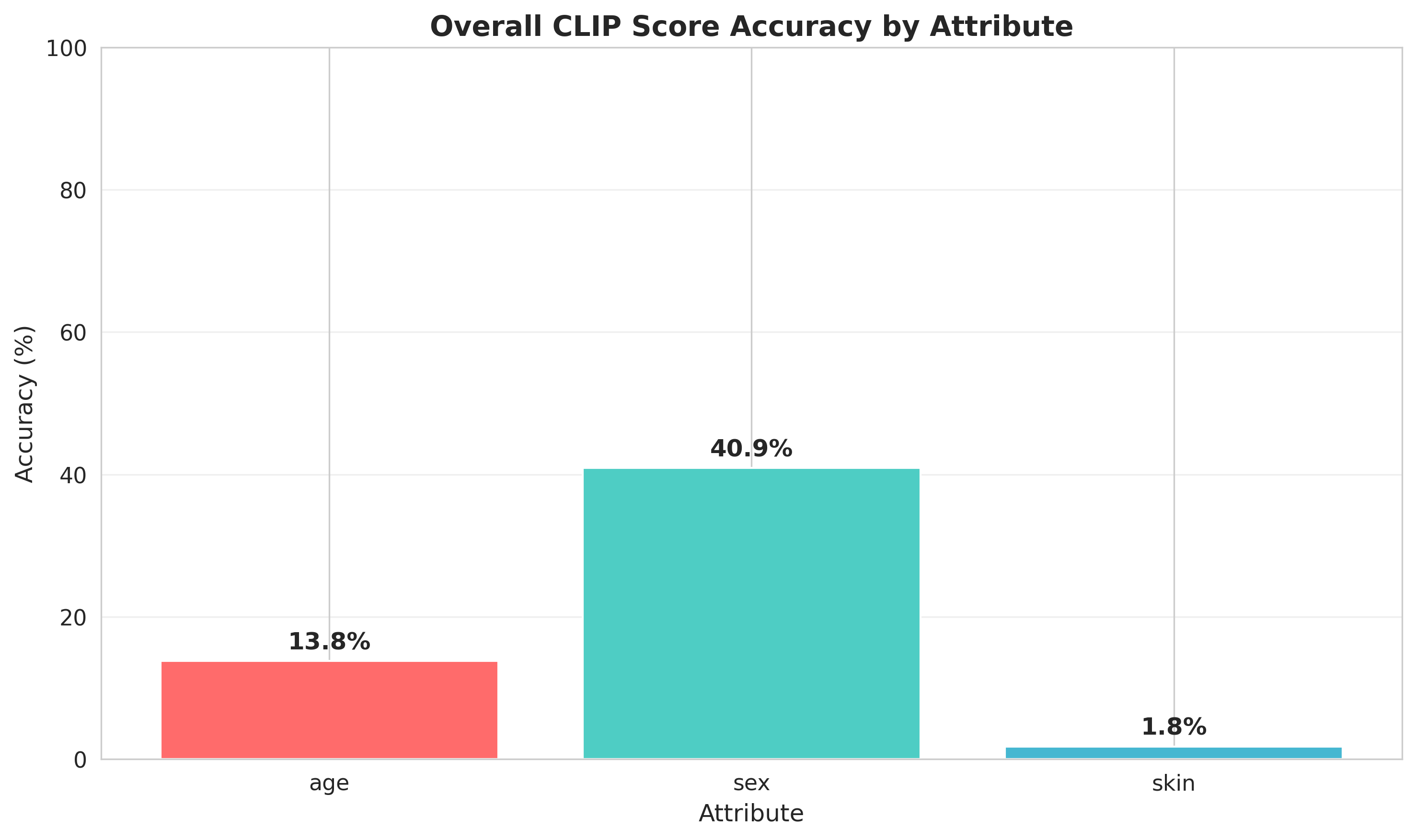}
        \caption{Overall CLIP Score Accuracy (real)}
    \end{subfigure}\hfill
    \begin{subfigure}{0.48\textwidth}
        \centering
        \includegraphics[width=\linewidth]{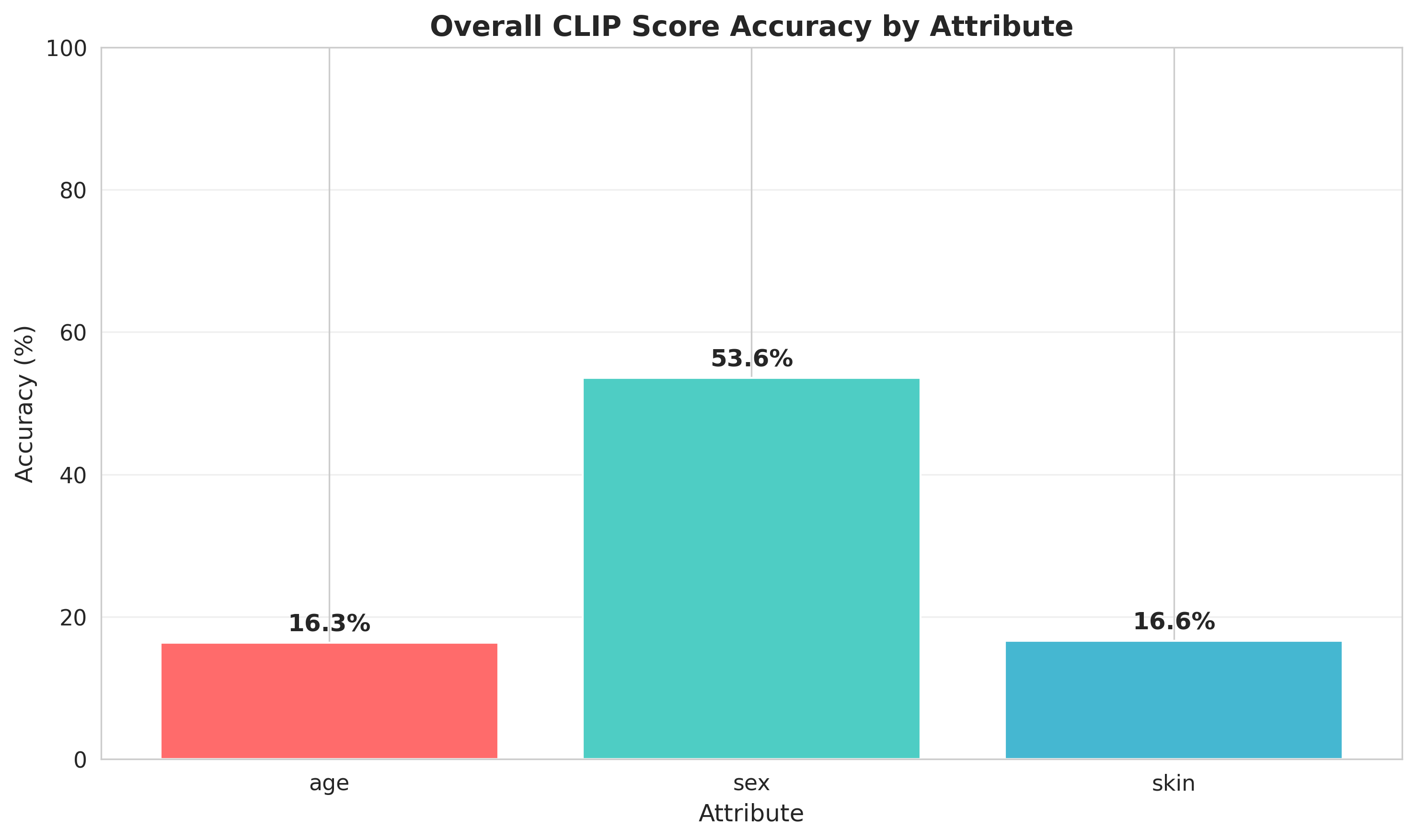}
        \caption{Overall CLIP Score Accuracy (synthetic)}
    \end{subfigure}

    \vspace{1em}

    \begin{subfigure}{0.48\textwidth}
        \centering
        \includegraphics[width=\linewidth]{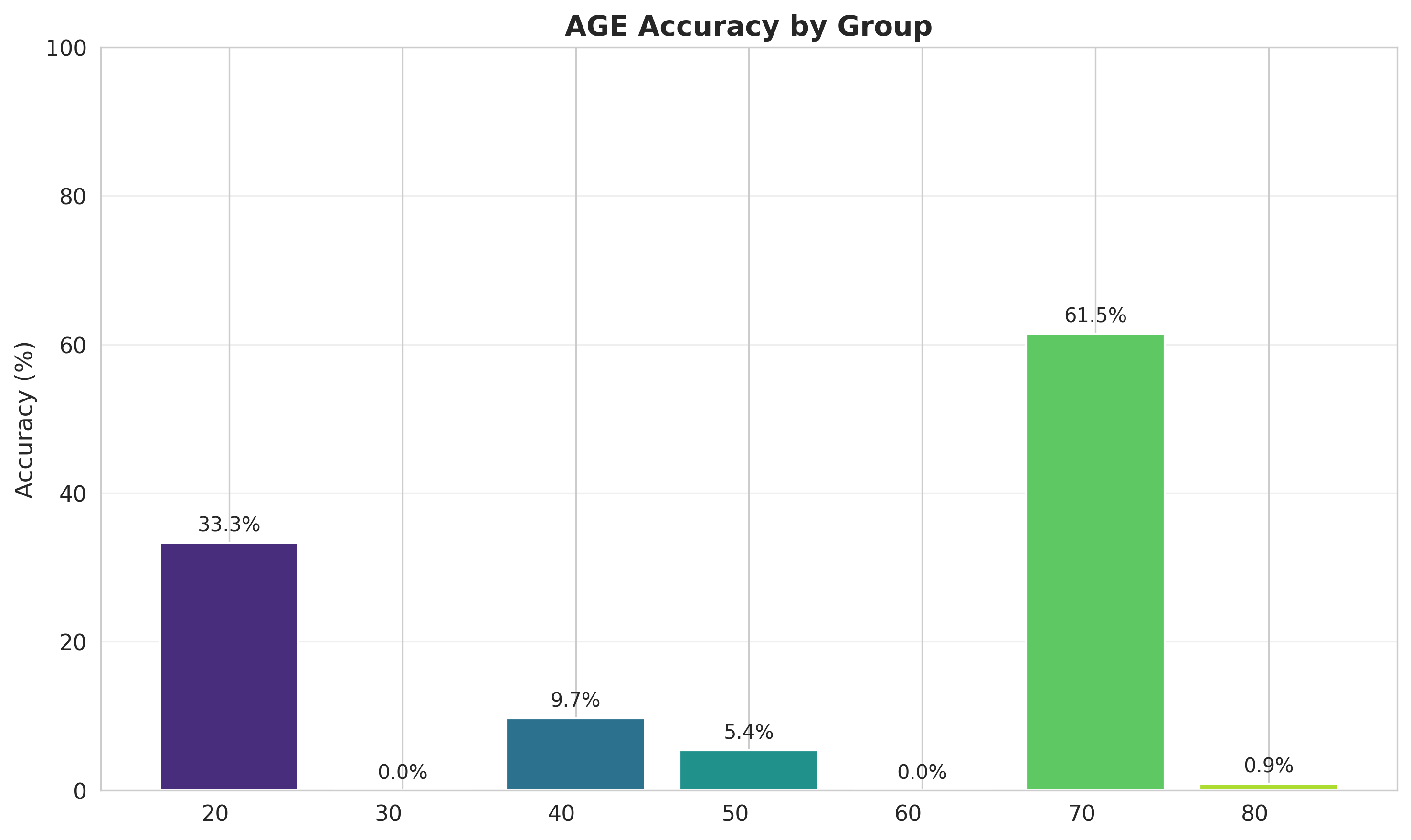}
        \caption{AGE Accuracy (real)}
    \end{subfigure}\hfill
    \begin{subfigure}{0.48\textwidth}
        \centering
        \includegraphics[width=\linewidth]{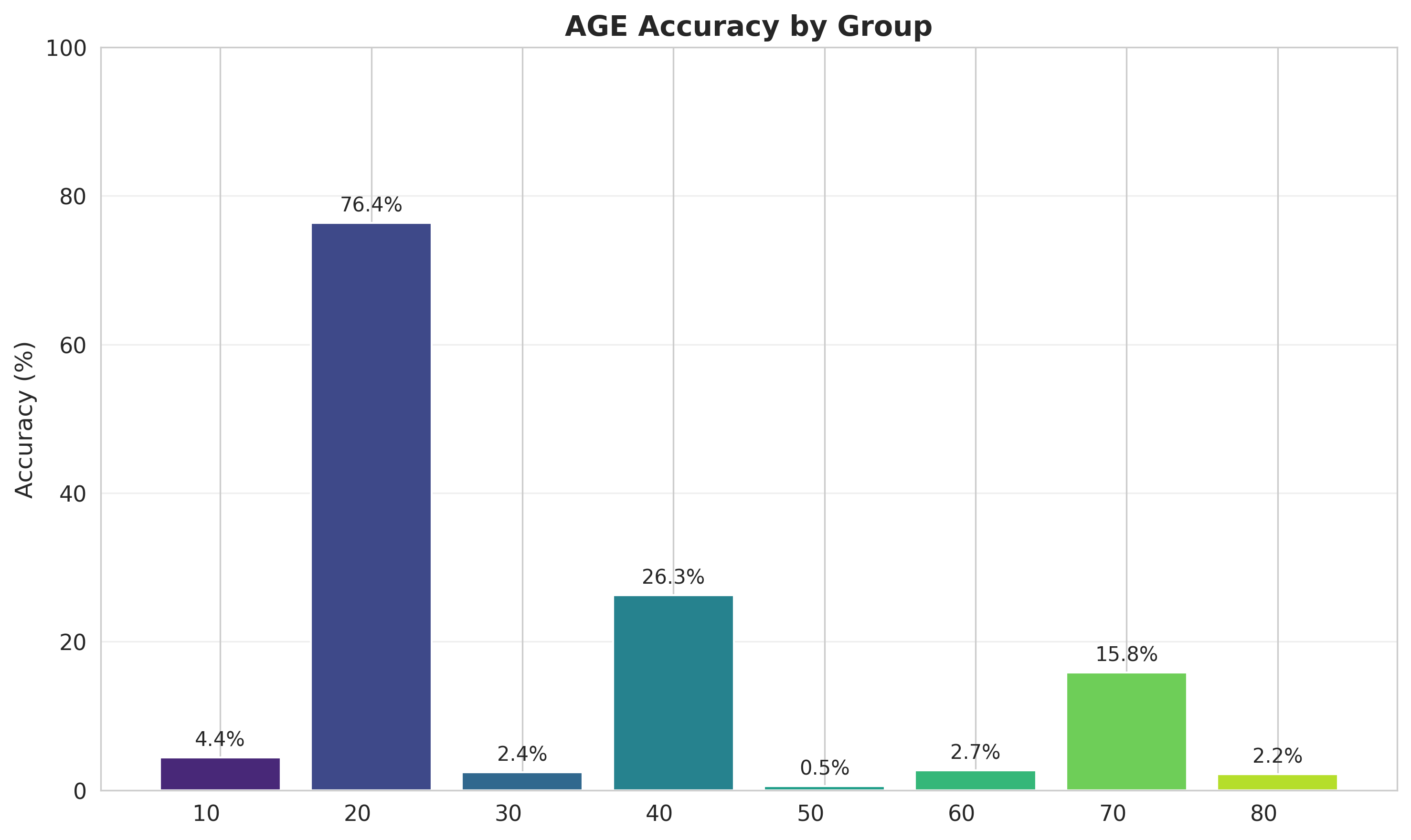}
        \caption{AGE Accuracy (synthetic)}
    \end{subfigure}

    \vspace{1em}

    \begin{subfigure}{0.48\textwidth}
        \centering
        \includegraphics[width=\linewidth]{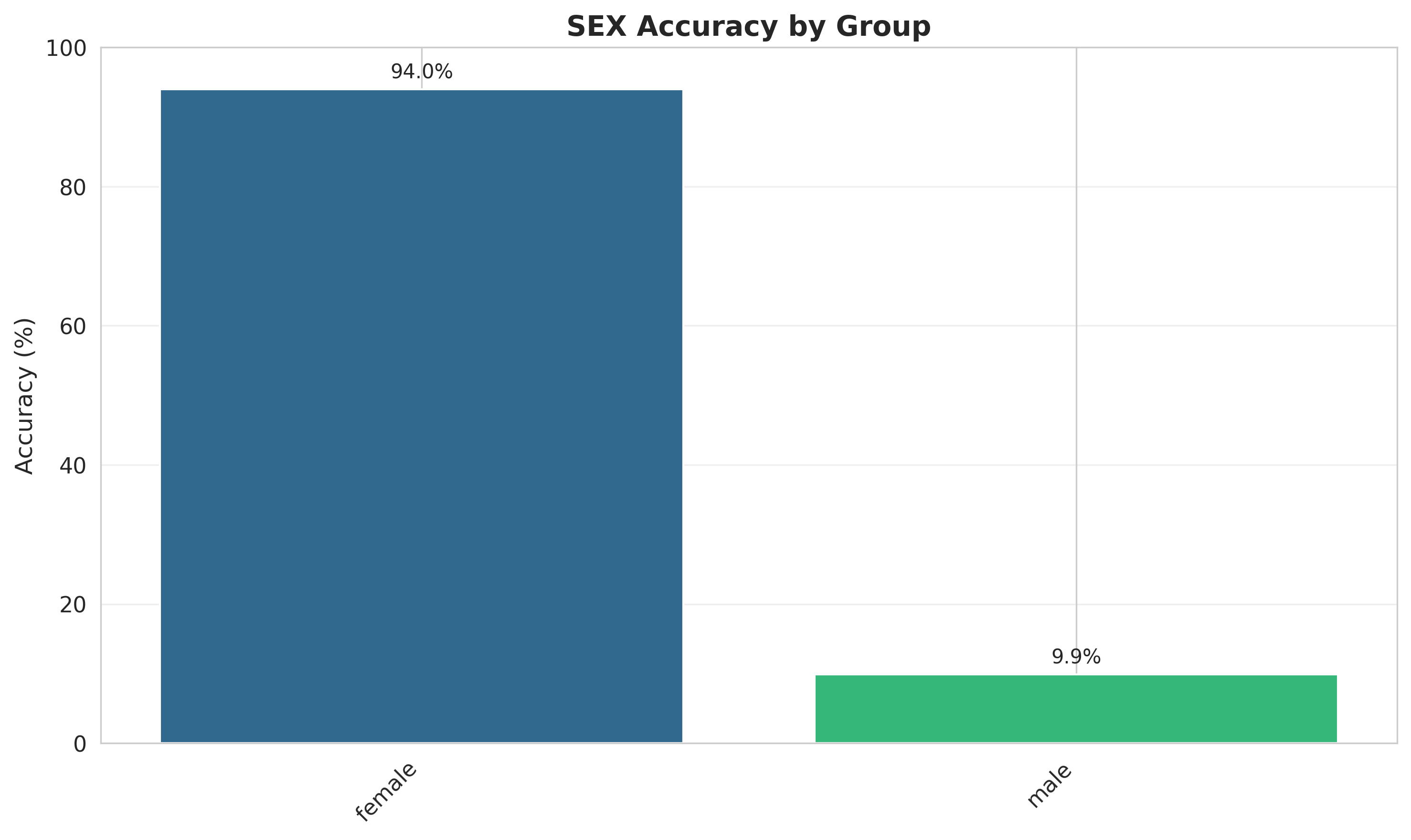}
        \caption{SEX Accuracy (real)}
    \end{subfigure}\hfill
    \begin{subfigure}{0.48\textwidth}
        \centering
        \includegraphics[width=\linewidth]{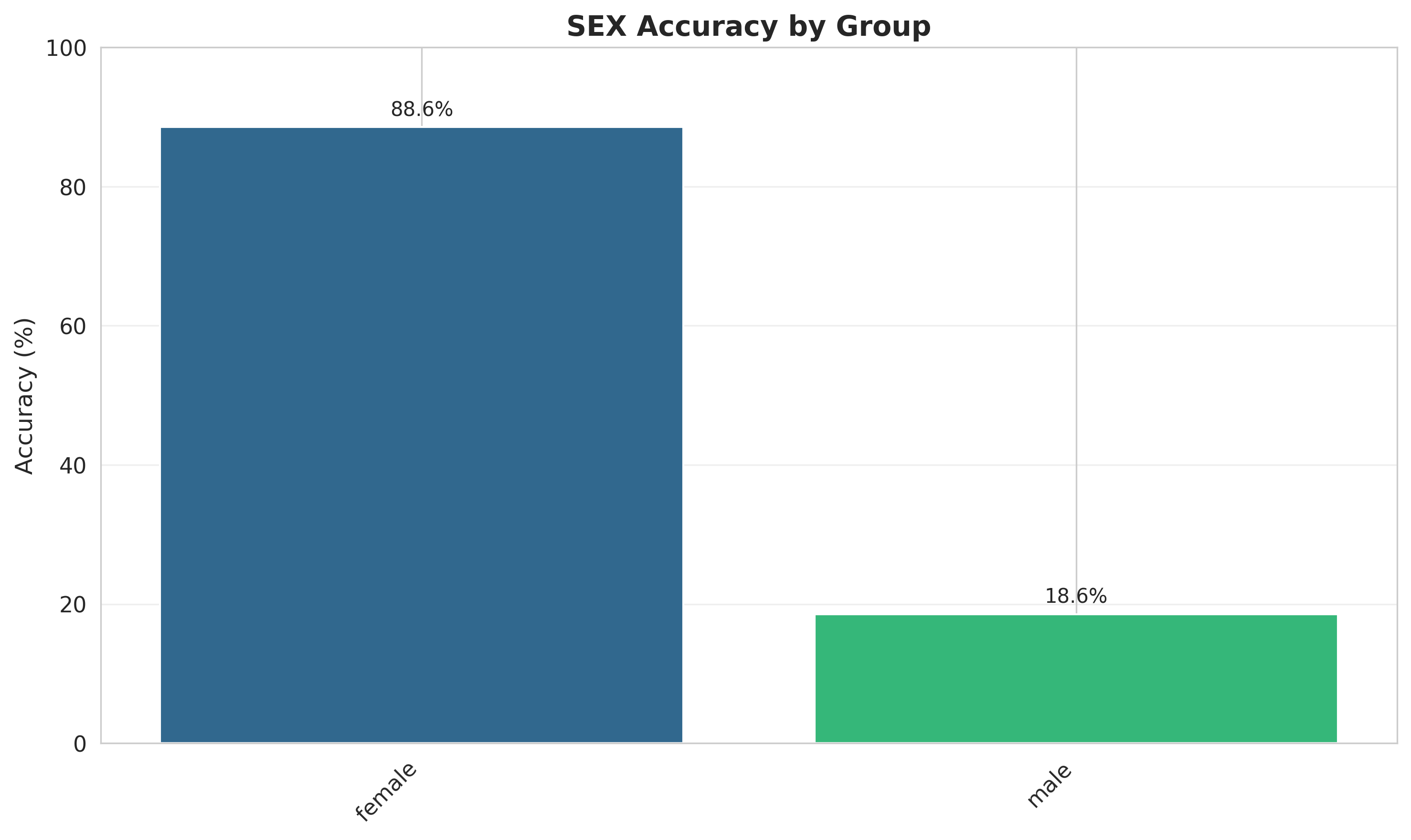}
        \caption{SEX Accuracy (synthetic)}
    \end{subfigure}

    \vspace{1em}

    \begin{subfigure}{0.48\textwidth}
        \centering
        \includegraphics[width=\linewidth]{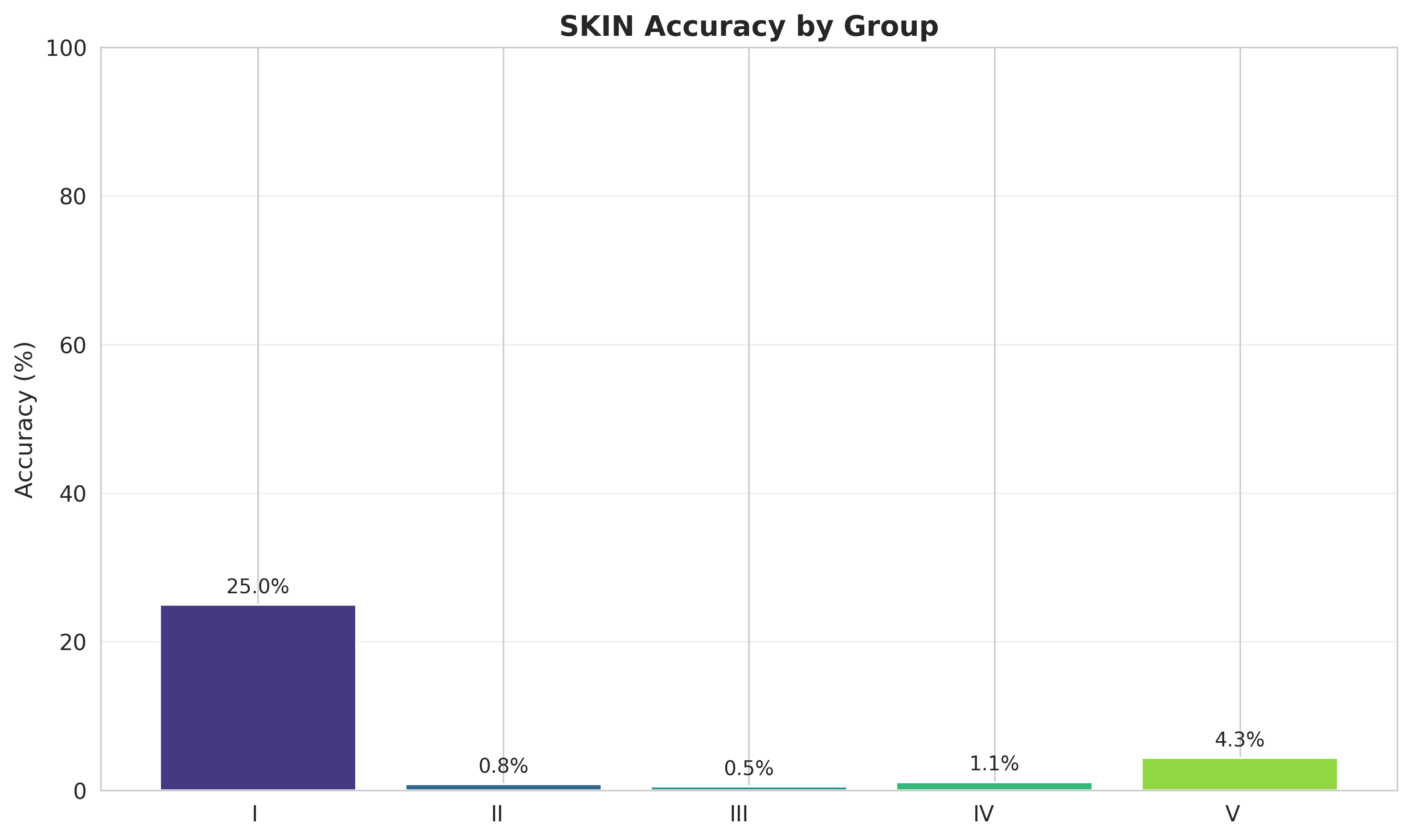}
        \caption{SKIN Accuracy (real)}
    \end{subfigure}\hfill
    \begin{subfigure}{0.48\textwidth}
        \centering
        \includegraphics[width=\linewidth]{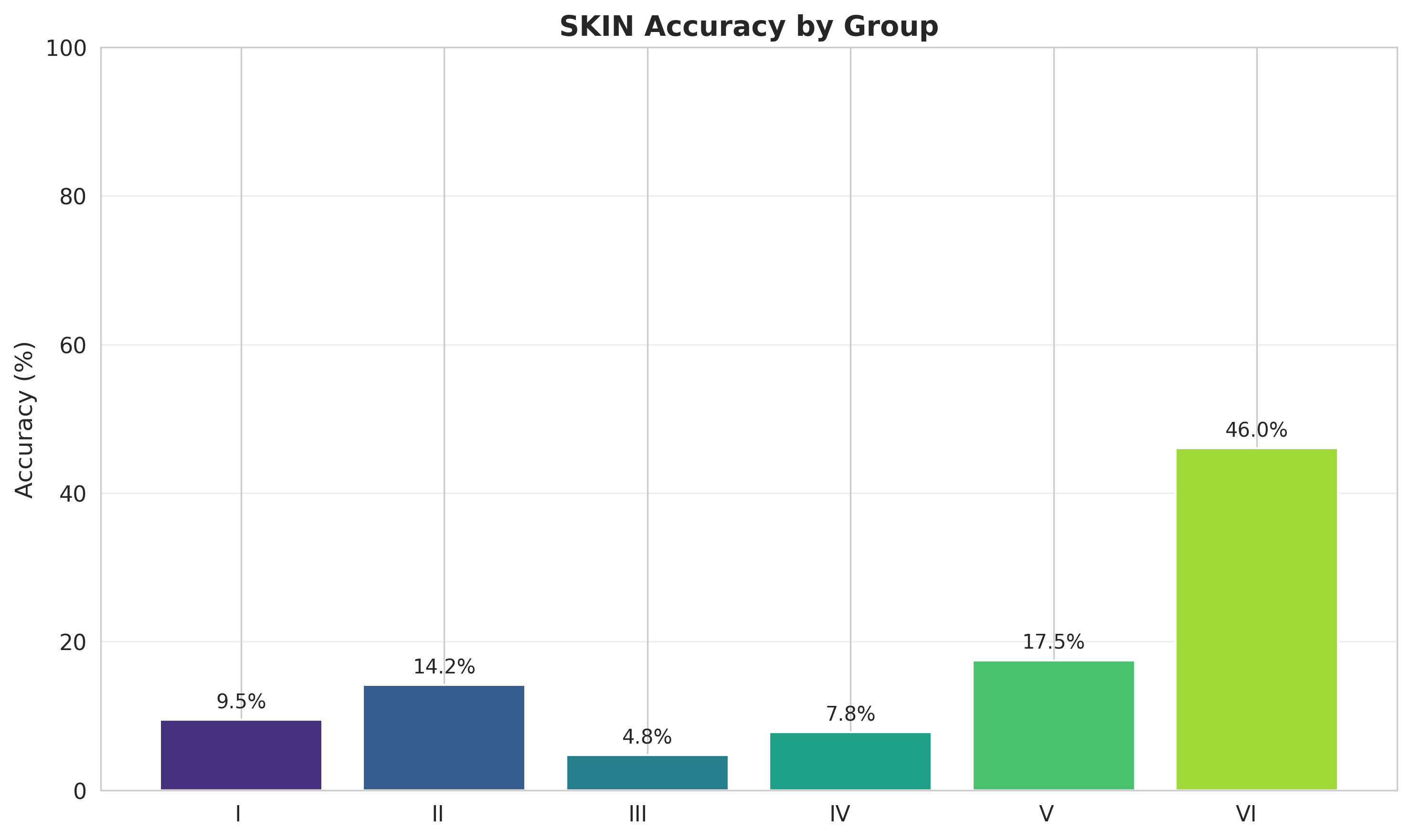}
        \caption{SKIN Accuracy (synthetic)}
    \end{subfigure}
    \caption{Comparison of CLIP-score classification accuracies across attributes.}
    \label{fig:clip_overall}
\end{figure*}

\begin{figure*}[t]
    \centering
    \includegraphics[width=\linewidth]{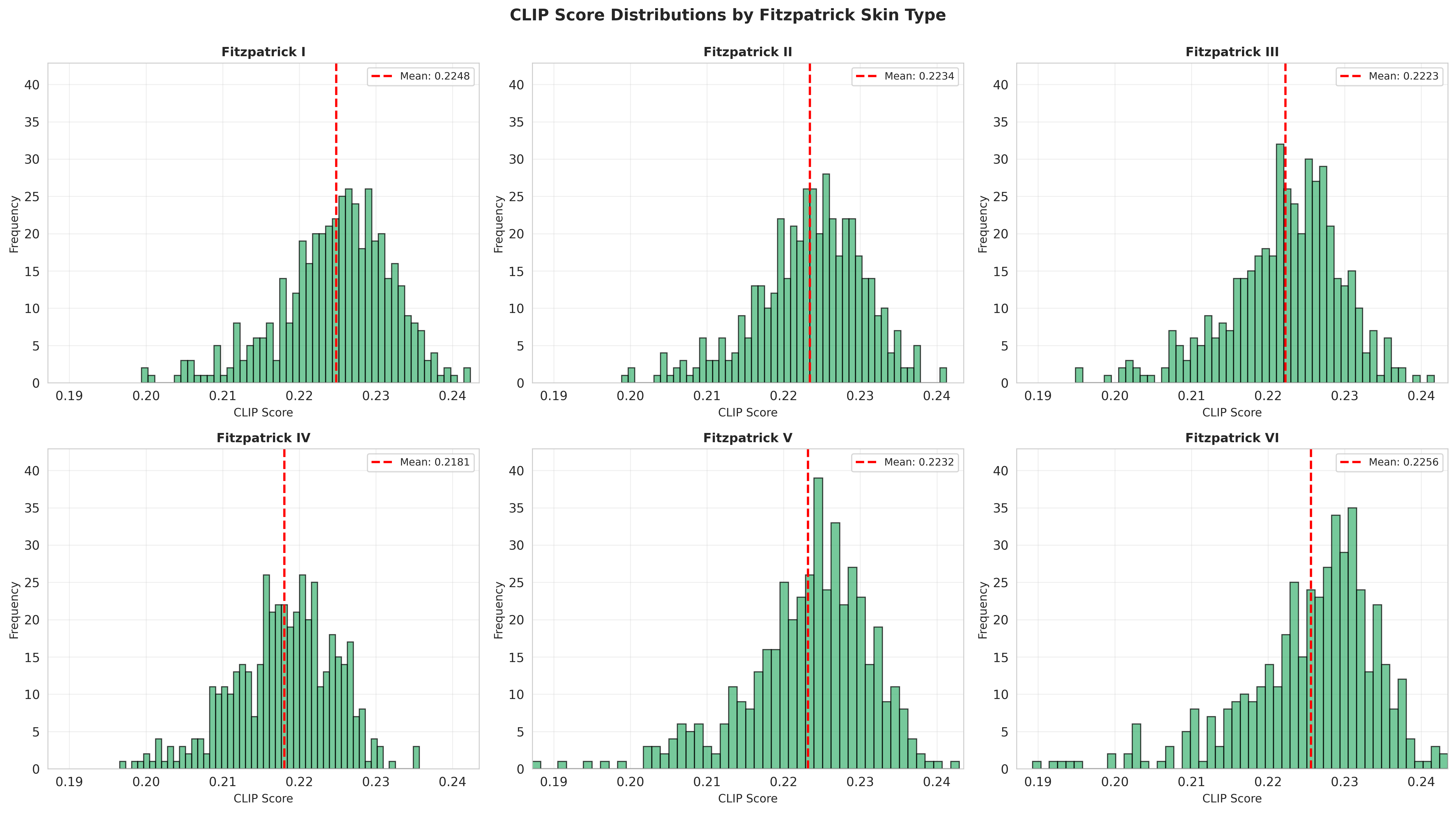}
    \caption{CLIP Score Distributions by Skin Type (real)}
    \vspace{1cm}
    \includegraphics[width=\linewidth]{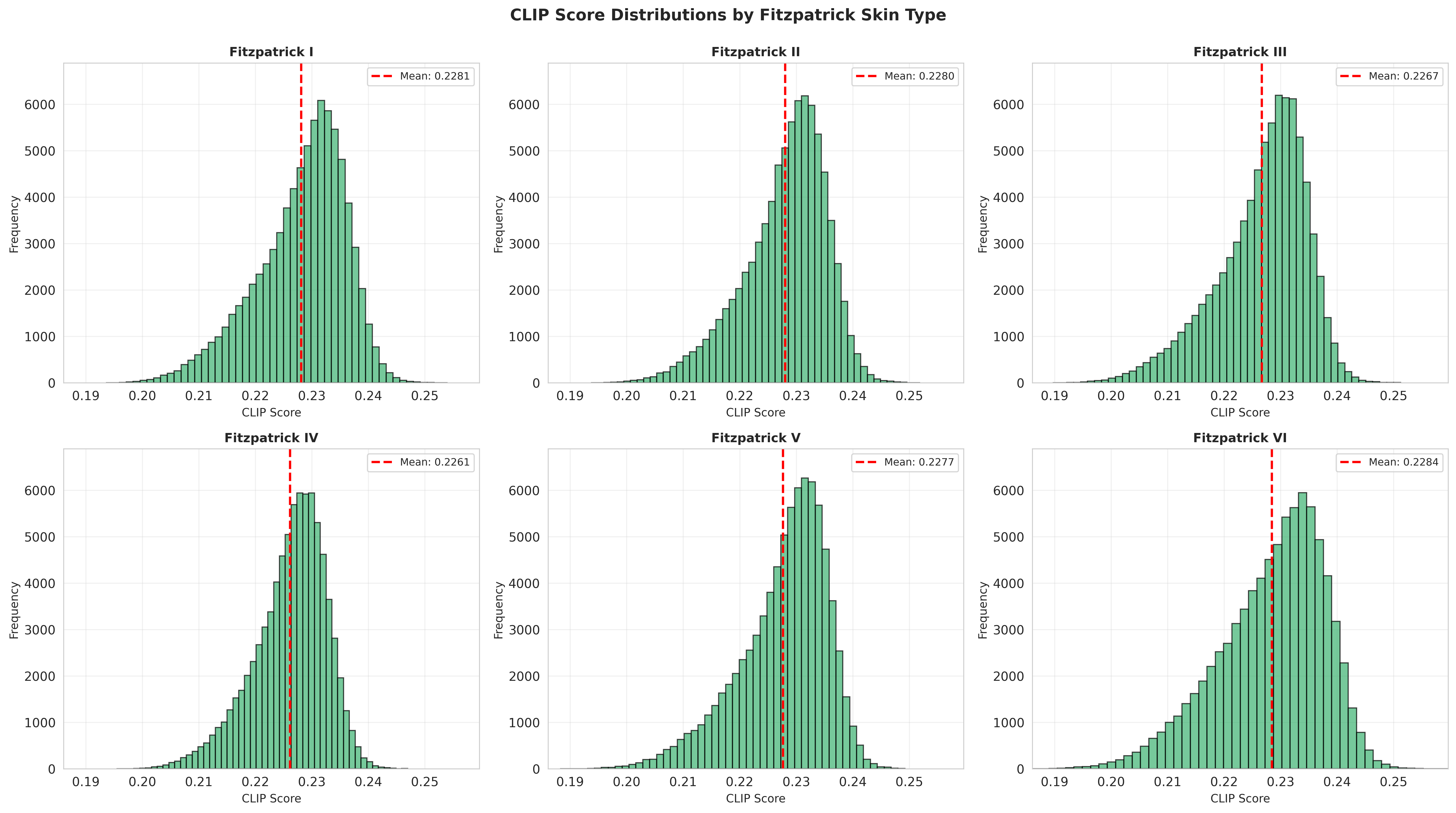}
    \caption{CLIP Score Distributions by Skin Type (synthetic)}
    \label{fig:clip_skin}
\end{figure*}

Finally, we are interested in the performance of WhyLesionCLIP \cite{yang2024textbook}, a fine-tuned version of OpenCLIP, on real as well as our synthetic images, see \autoref{fig:clip_overall} to \autoref{fig:clip_skin}.
WhyLesionCLIP \cite{yang2024textbook} captures dermatological and clinical semantics more reliably than general purpose CLIP models and is thus a good choice for general assessment.
The zero-shot predictions were obtained using three prompt families corresponding to age, sex, and Fitzpatrick skin type.
Age prompts covered the eight age groups and followed the template
``A clinical close-up photograph of the \{body\_site\} skin of a \{age\}-year-old melanoma patient.'' 
Sex prompts described the patient's sex using the format 
``A clinical close-up photograph of the \{body\_site\} skin of a \{sex\} melanoma patient.'' 
Fitzpatrick skin type prompts covered types~I through~VI using the template 
``A clinical close-up of the skin of a melanoma patient with Fitzpatrick type~\{type\} (\{description of the skin type\}).''
Overall, our synthetic images follow similar performance pattern across the age-sex-skin attributes, but our synthetic images are more recognizable as indicated by higher accuracies for all attribute groups.
Interestingly, the performance across age groups is very imbalanced and varies a lot between real and synthetic showing that recognizing age from skin lesion images alone is difficult.
Furthermore, both real and synthetic images are highly imbalances w.r.t to sex classification.
Male accuracy is significantly lower than female accuracy, but our synthetic images are much better than real ones while reaching comparable performance for female.
In terms of skin type classification, our synthetic images reach higher performance on all but skin type I, which is the best in real data.
Finally, we visualize the CLIP score distributions by age, sex and skin type.
Our synthetic images consistently lead to smoother distributions with higher mean values indicating better recognizability.

\begin{figure*}[t]
\centering
\vspace{1em}

\resizebox{\textwidth}{!}{%
\setlength{\tabcolsep}{2pt} 

\begin{tabular}{
    c
    *{2}{c}@{\hspace{2pt}}  
    *{2}{c}@{\hspace{2pt}}  
    *{2}{c}@{\hspace{2pt}}  
    *{2}{c}@{\hspace{2pt}}  
    *{2}{c}@{\hspace{2pt}}  
    *{2}{c}                 
}
\rowlabel{Age 10} &
\samplesynthimg{10}{female}{I} & \samplesynthimg{10}{male}{I} &
\samplesynthimg{10}{female}{II} & \samplesynthimg{10}{male}{II} &
\samplesynthimg{10}{female}{III} & \samplesynthimg{10}{male}{III} &
\samplesynthimg{10}{female}{IV} & \samplesynthimg{10}{male}{IV} &
\samplesynthimg{10}{female}{V} & \samplesynthimg{10}{male}{V} &
\samplesynthimg{10}{female}{VI} & \samplesynthimg{10}{male}{VI} \\

\rowlabel{Age 20} &
\samplesynthimg{20}{female}{I} & \samplesynthimg{20}{male}{I} &
\samplesynthimg{20}{female}{II} & \samplesynthimg{20}{male}{II} &
\samplesynthimg{20}{female}{III} & \samplesynthimg{20}{male}{III} &
\samplesynthimg{20}{female}{IV} & \samplesynthimg{20}{male}{IV} &
\samplesynthimg{20}{female}{V} & \samplesynthimg{20}{male}{V} &
\samplesynthimg{20}{female}{VI} & \samplesynthimg{20}{male}{VI} \\

\rowlabel{Age 30} &
\samplesynthimg{30}{female}{I} & \samplesynthimg{30}{male}{I} &
\samplesynthimg{30}{female}{II} & \samplesynthimg{30}{male}{II} &
\samplesynthimg{30}{female}{III} & \samplesynthimg{30}{male}{III} &
\samplesynthimg{30}{female}{IV} & \samplesynthimg{30}{male}{IV} &
\samplesynthimg{30}{female}{V} & \samplesynthimg{30}{male}{V} &
\samplesynthimg{30}{female}{VI} & \samplesynthimg{30}{male}{VI} \\

\rowlabel{Age 40} &
\samplesynthimg{40}{female}{I} & \samplesynthimg{40}{male}{I} &
\samplesynthimg{40}{female}{II} & \samplesynthimg{40}{male}{II} &
\samplesynthimg{40}{female}{III} & \samplesynthimg{40}{male}{III} &
\samplesynthimg{40}{female}{IV} & \samplesynthimg{40}{male}{IV} &
\samplesynthimg{40}{female}{V} & \samplesynthimg{40}{male}{V} &
\samplesynthimg{40}{female}{VI} & \samplesynthimg{40}{male}{VI} \\

\rowlabel{Age 50} &
\samplesynthimg{50}{female}{I} & \samplesynthimg{50}{male}{I} &
\samplesynthimg{50}{female}{II} & \samplesynthimg{50}{male}{II} &
\samplesynthimg{50}{female}{III} & \samplesynthimg{50}{male}{III} &
\samplesynthimg{50}{female}{IV} & \samplesynthimg{50}{male}{IV} &
\samplesynthimg{50}{female}{V} & \samplesynthimg{50}{male}{V} &
\samplesynthimg{50}{female}{VI} & \samplesynthimg{50}{male}{VI} \\

\rowlabel{Age 60} &
\samplesynthimg{60}{female}{I} & \samplesynthimg{60}{male}{I} &
\samplesynthimg{60}{female}{II} & \samplesynthimg{60}{male}{II} &
\samplesynthimg{60}{female}{III} & \samplesynthimg{60}{male}{III} &
\samplesynthimg{60}{female}{IV} & \samplesynthimg{60}{male}{IV} &
\samplesynthimg{60}{female}{V} & \samplesynthimg{60}{male}{V} &
\samplesynthimg{60}{female}{VI} & \samplesynthimg{60}{male}{VI} \\

\rowlabel{Age 70} &
\samplesynthimg{70}{female}{I} & \samplesynthimg{70}{male}{I} &
\samplesynthimg{70}{female}{II} & \samplesynthimg{70}{male}{II} &
\samplesynthimg{70}{female}{III} & \samplesynthimg{70}{male}{III} &
\samplesynthimg{70}{female}{IV} & \samplesynthimg{70}{male}{IV} &
\samplesynthimg{70}{female}{V} & \samplesynthimg{70}{male}{V} &
\samplesynthimg{70}{female}{VI} & \samplesynthimg{70}{male}{VI} \\

\rowlabel{Age 80} &
\samplesynthimg{80}{female}{I} & \samplesynthimg{80}{male}{I} &
\samplesynthimg{80}{female}{II} & \samplesynthimg{80}{male}{II} &
\samplesynthimg{80}{female}{III} & \samplesynthimg{80}{male}{III} &
\samplesynthimg{80}{female}{IV} & \samplesynthimg{80}{male}{IV} &
\samplesynthimg{80}{female}{V} & \samplesynthimg{80}{male}{V} &
\samplesynthimg{80}{female}{VI} & \samplesynthimg{80}{male}{VI} \\

&
\colheading{Skin Type I}{Female} &
\colheading{Skin Type I}{Male} &
\colheading{Skin Type II}{Female} &
\colheading{Skin Type II}{Male} &
\colheading{Skin Type III}{Female} &
\colheading{Skin Type III}{Male} &
\colheading{Skin Type IV}{Female} &
\colheading{Skin Type IV}{Male} &
\colheading{Skin Type V}{Female} &
\colheading{Skin Type V}{Male} &
\colheading{Skin Type VI}{Female} &
\colheading{Skin Type VI}{Male} \\

\end{tabular}
} 
\caption{Synthetic melanoma images generated by our model with default sampling parameters (cfg=8, steps=200). Rows represent Fitzpatrick skin types (I--VI) combined with sex, and columns represent age groups (10--80). The grid demonstrates coverage of diverse demographic groups for fairness assessment.}
  \label{fig:synth_melanoma_samples}
\end{figure*}

\begin{figure*}[t]
\centering
\vspace{1em}

\resizebox{\textwidth}{!}{%
\setlength{\tabcolsep}{2pt} 

\begin{tabular}{
    c
    *{2}{c}@{\hspace{2pt}}  
    *{2}{c}@{\hspace{2pt}}  
    *{2}{c}@{\hspace{2pt}}  
    *{2}{c}@{\hspace{2pt}}  
    *{2}{c}@{\hspace{2pt}}  
    *{2}{c}                 
}
\rowlabel{Age 10} &
\sampleimg{10}{female}{I} & \sampleimg{10}{male}{I} &
\sampleimg{10}{female}{II} & \sampleimg{10}{male}{II} &
\sampleimg{10}{female}{III} & \sampleimg{10}{male}{III} &
\sampleimg{10}{female}{IV} & \sampleimg{10}{male}{IV} &
\sampleimg{10}{female}{V} & \sampleimg{10}{male}{V} &
\sampleimg{10}{female}{VI} & \sampleimg{10}{male}{VI} \\

\rowlabel{Age 20} &
\sampleimg{20}{female}{I} & \sampleimg{20}{male}{I} &
\sampleimg{20}{female}{II} & \sampleimg{20}{male}{II} &
\sampleimg{20}{female}{III} & \sampleimg{20}{male}{III} &
\sampleimg{20}{female}{IV} & \sampleimg{20}{male}{IV} &
\sampleimg{20}{female}{V} & \sampleimg{20}{male}{V} &
\sampleimg{20}{female}{VI} & \sampleimg{20}{male}{VI} \\

\rowlabel{Age 30} &
\sampleimg{30}{female}{I} & \sampleimg{30}{male}{I} &
\sampleimg{30}{female}{II} & \sampleimg{30}{male}{II} &
\sampleimg{30}{female}{III} & \sampleimg{30}{male}{III} &
\sampleimg{30}{female}{IV} & \sampleimg{30}{male}{IV} &
\sampleimg{30}{female}{V} & \sampleimg{30}{male}{V} &
\sampleimg{30}{female}{VI} & \sampleimg{30}{male}{VI} \\

\rowlabel{Age 40} &
\sampleimg{40}{female}{I} & \sampleimg{40}{male}{I} &
\sampleimg{40}{female}{II} & \sampleimg{40}{male}{II} &
\sampleimg{40}{female}{III} & \sampleimg{40}{male}{III} &
\sampleimg{40}{female}{IV} & \sampleimg{40}{male}{IV} &
\sampleimg{40}{female}{V} & \sampleimg{40}{male}{V} &
\sampleimg{40}{female}{VI} & \sampleimg{40}{male}{VI} \\

\rowlabel{Age 50} &
\sampleimg{50}{female}{I} & \sampleimg{50}{male}{I} &
\sampleimg{50}{female}{II} & \sampleimg{50}{male}{II} &
\sampleimg{50}{female}{III} & \sampleimg{50}{male}{III} &
\sampleimg{50}{female}{IV} & \sampleimg{50}{male}{IV} &
\sampleimg{50}{female}{V} & \sampleimg{50}{male}{V} &
\sampleimg{50}{female}{VI} & \sampleimg{50}{male}{VI} \\

\rowlabel{Age 60} &
\sampleimg{60}{female}{I} & \sampleimg{60}{male}{I} &
\sampleimg{60}{female}{II} & \sampleimg{60}{male}{II} &
\sampleimg{60}{female}{III} & \sampleimg{60}{male}{III} &
\sampleimg{60}{female}{IV} & \sampleimg{60}{male}{IV} &
\sampleimg{60}{female}{V} & \sampleimg{60}{male}{V} &
\sampleimg{60}{female}{VI} & \sampleimg{60}{male}{VI} \\

\rowlabel{Age 70} &
\sampleimg{70}{female}{I} & \sampleimg{70}{male}{I} &
\sampleimg{70}{female}{II} & \sampleimg{70}{male}{II} &
\sampleimg{70}{female}{III} & \sampleimg{70}{male}{III} &
\sampleimg{70}{female}{IV} & \sampleimg{70}{male}{IV} &
\sampleimg{70}{female}{V} & \sampleimg{70}{male}{V} &
\sampleimg{70}{female}{VI} & \sampleimg{70}{male}{VI} \\

\rowlabel{Age 80} &
\sampleimg{80}{female}{I} & \sampleimg{80}{male}{I} &
\sampleimg{80}{female}{II} & \sampleimg{80}{male}{II} &
\sampleimg{80}{female}{III} & \sampleimg{80}{male}{III} &
\sampleimg{80}{female}{IV} & \sampleimg{80}{male}{IV} &
\sampleimg{80}{female}{V} & \sampleimg{80}{male}{V} &
\sampleimg{80}{female}{VI} & \sampleimg{80}{male}{VI} \\

&
\colheading{Skin Type I}{Female} &
\colheading{Skin Type I}{Male} &
\colheading{Skin Type II}{Female} &
\colheading{Skin Type II}{Male} &
\colheading{Skin Type III}{Female} &
\colheading{Skin Type III}{Male} &
\colheading{Skin Type IV}{Female} &
\colheading{Skin Type IV}{Male} &
\colheading{Skin Type V}{Female} &
\colheading{Skin Type V}{Male} &
\colheading{Skin Type VI}{Female} &
\colheading{Skin Type VI}{Male} \\

\end{tabular}
} 
\caption{Real melanoma images from the MILK10k dataset \cite{MILK10k_2025}. Rows represent Fitzpatrick skin types (I--VI) combined with sex, and columns represent age groups (10--80). The grid demonstrates coverage of diverse demographic groups for fairness assessment. Many age–skin–sex combinations are missing from the real dataset, making fairness assessment difficult.}
  \label{fig:melanoma_samples}
\end{figure*}

\begin{table*}[t]
\centering
\begin{minipage}{0.55\linewidth}
\centering
\begin{tabular}{c cccccc}
\hline
cfg $\backslash$ steps & 100 & 150 & 200 & 250 & 300 & 400 \\
\hline
4.0  & 29.69 & 28.64 & 28.23 & 32.55 & 27.93 & \textbf{27.77} \\
6.7  & 30.66 & 30.00 & 29.75 & 32.37 & 29.61 & 29.57 \\
8.0  & 31.59 & 30.92 & \underline{30.75} & 32.57 & 30.71 & 30.73 \\
10.0 & 32.53 & 32.34 & 32.38 & 34.15 & 32.47 & 32.52 \\
12.0 & 34.38 & 34.42 & 34.50 & 34.14 & 34.79 & 34.90 \\
\hline
\end{tabular}
\caption{FID averaged over age groups for each combination of
classifier-free guidance (cfg) and diffusion steps showing that default setting (underlined) are not optimal settings (bold) when trained on skin lesions.}
\label{tab:fid_cfg_steps_ageavg}
\end{minipage}
\hfill
\begin{minipage}{0.35\linewidth}
\centering
\begin{tabular}{l c c}
\hline
Age & Mean FID & Data ratio\\
\hline
10 & 40.56 & 0.35\% \\
20 & 36.88 & 0.66\% \\
30 & 36.05 & 3.13\% \\
40 & 30.84 & 8.06\% \\
50 & 26.08 & 11.57\% \\
60 & \textbf{25.74} & 12.50\% \\
70 & 27.60 & 9.37\% \\
80 & 28.94 & 5.00\% \\
\hline
\end{tabular}
\caption{Data ratio and mean FID over all cfg and steps for each age group. Generation quality generally correlates with the available training data.}
\label{tab:fid_age_means}
\end{minipage}
\end{table*}

\begin{figure*}[t]
    \centering
    \includegraphics[width=\linewidth]{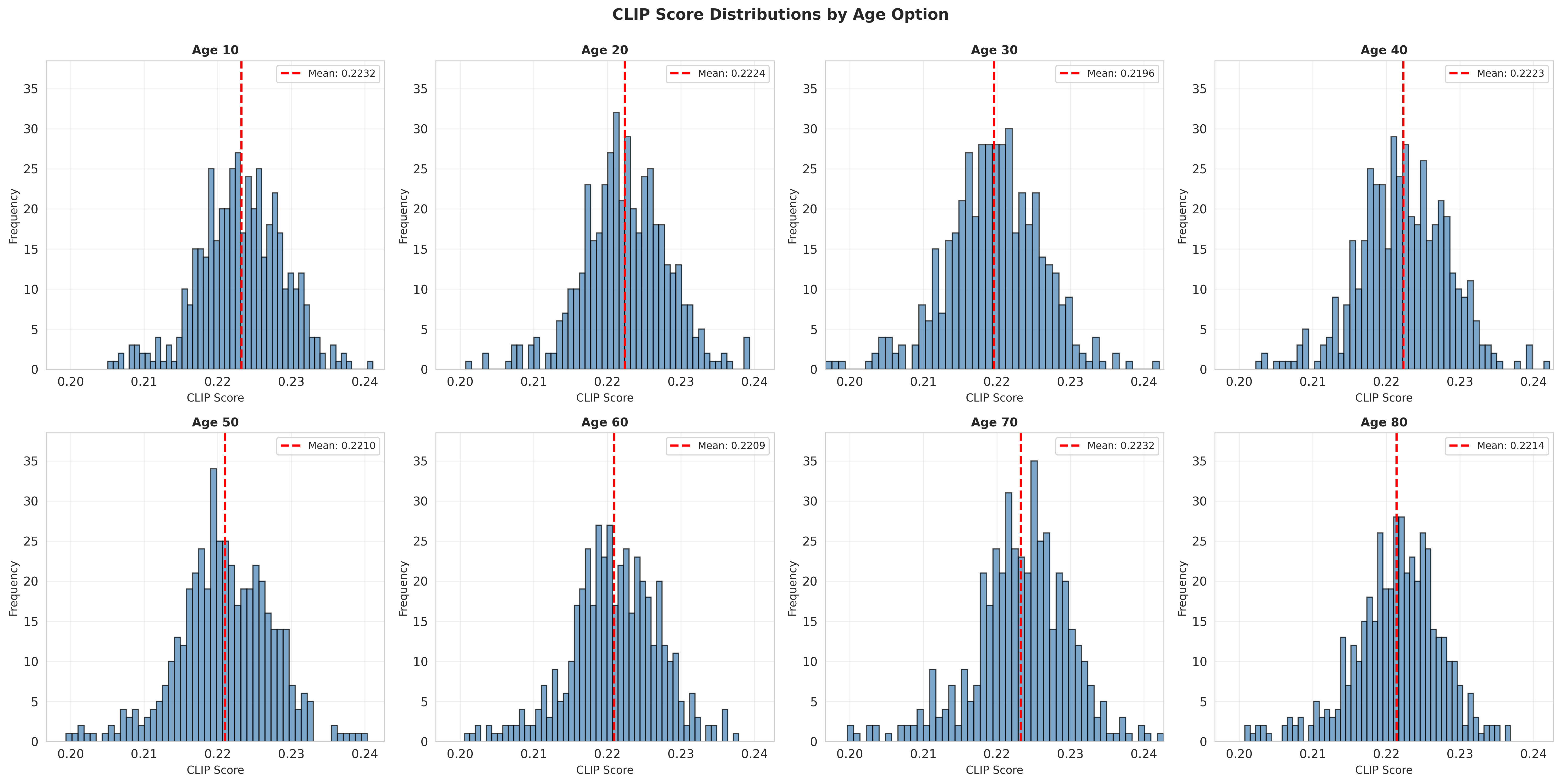}
    \caption{CLIP Score Distributions by Age (real)}
    \vspace{1cm}
    \includegraphics[width=\linewidth]{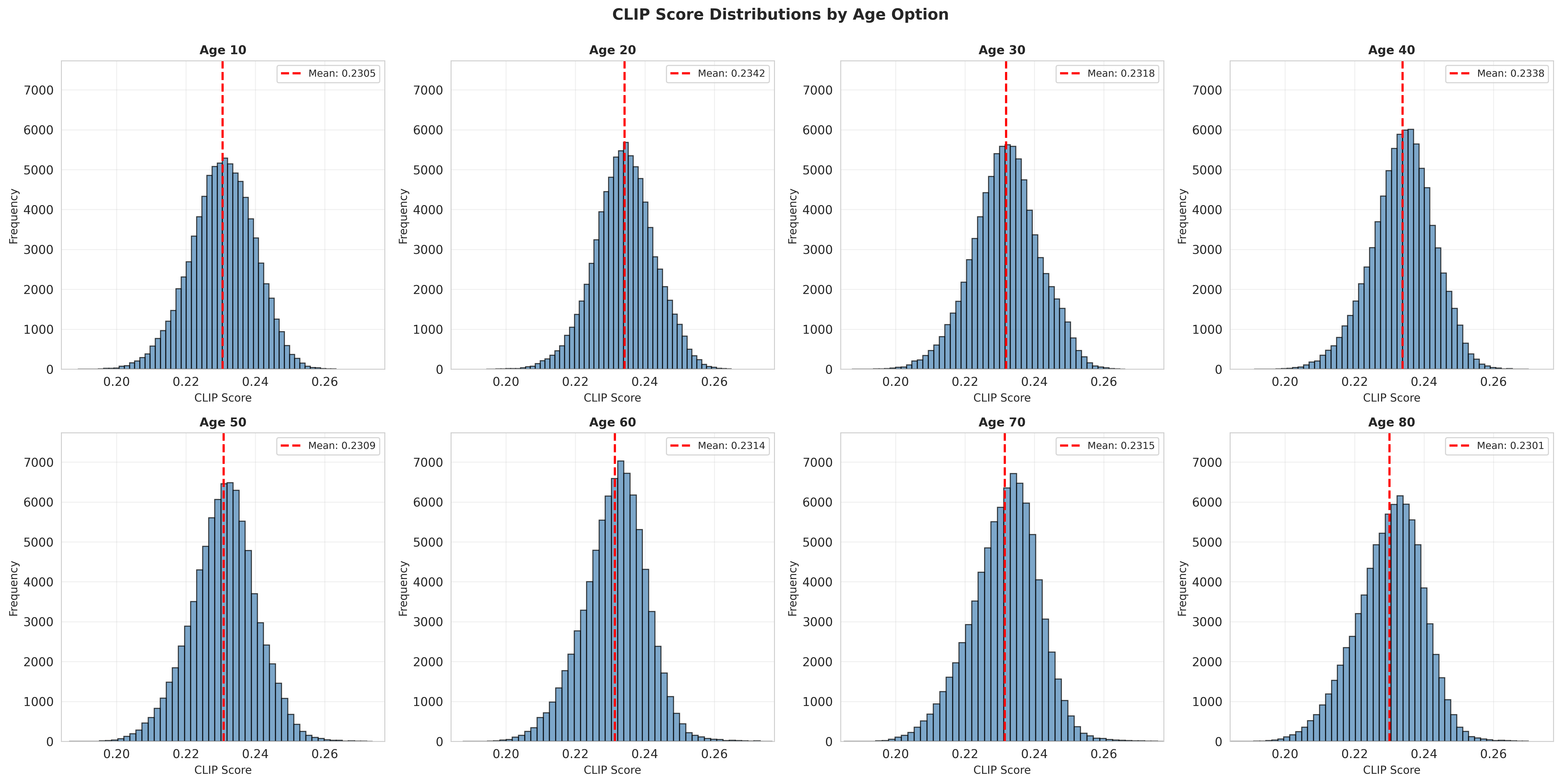}
    \caption{CLIP Score Distributions by Age (synthetic)}
\end{figure*}

\begin{figure*}[t]
    \centering
    \includegraphics[width=\linewidth]{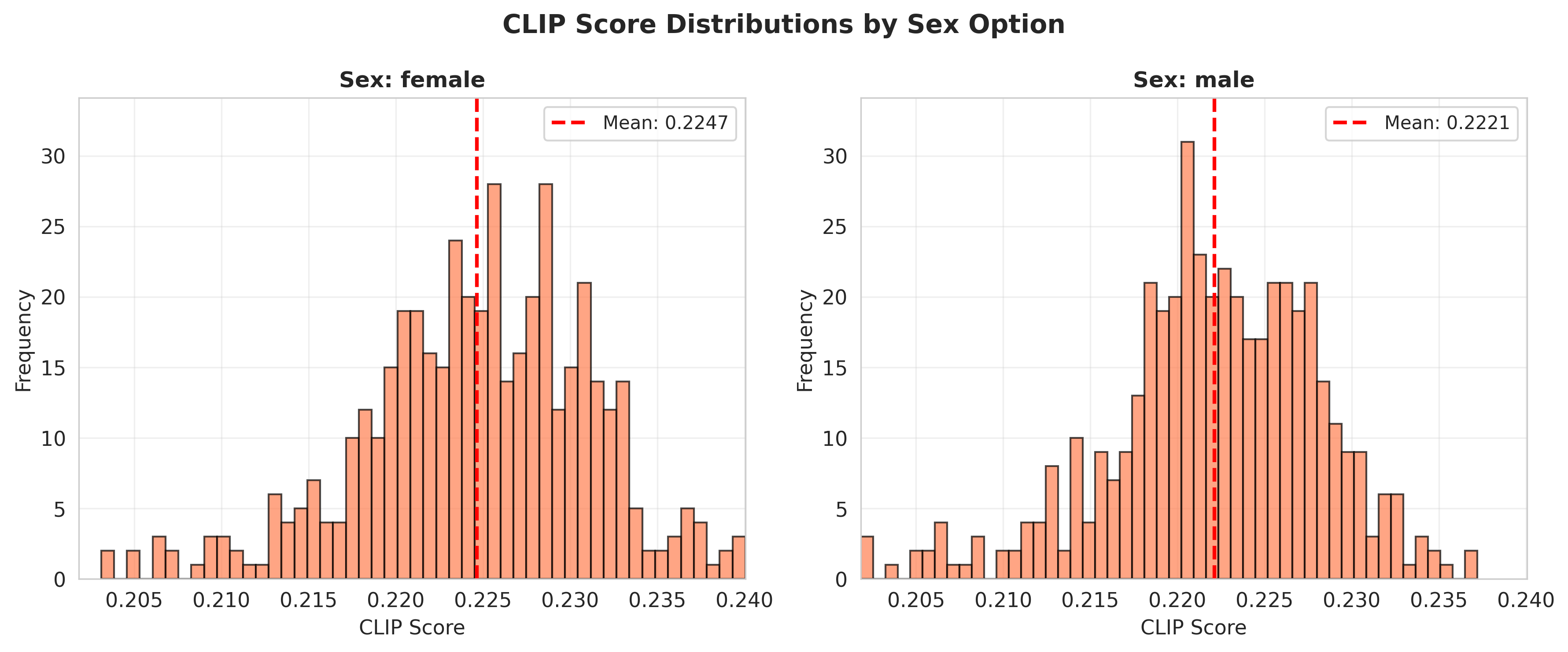}
    \caption{CLIP Score Distributions by Sex (real)}
    \vspace{1cm}
    \includegraphics[width=\linewidth]{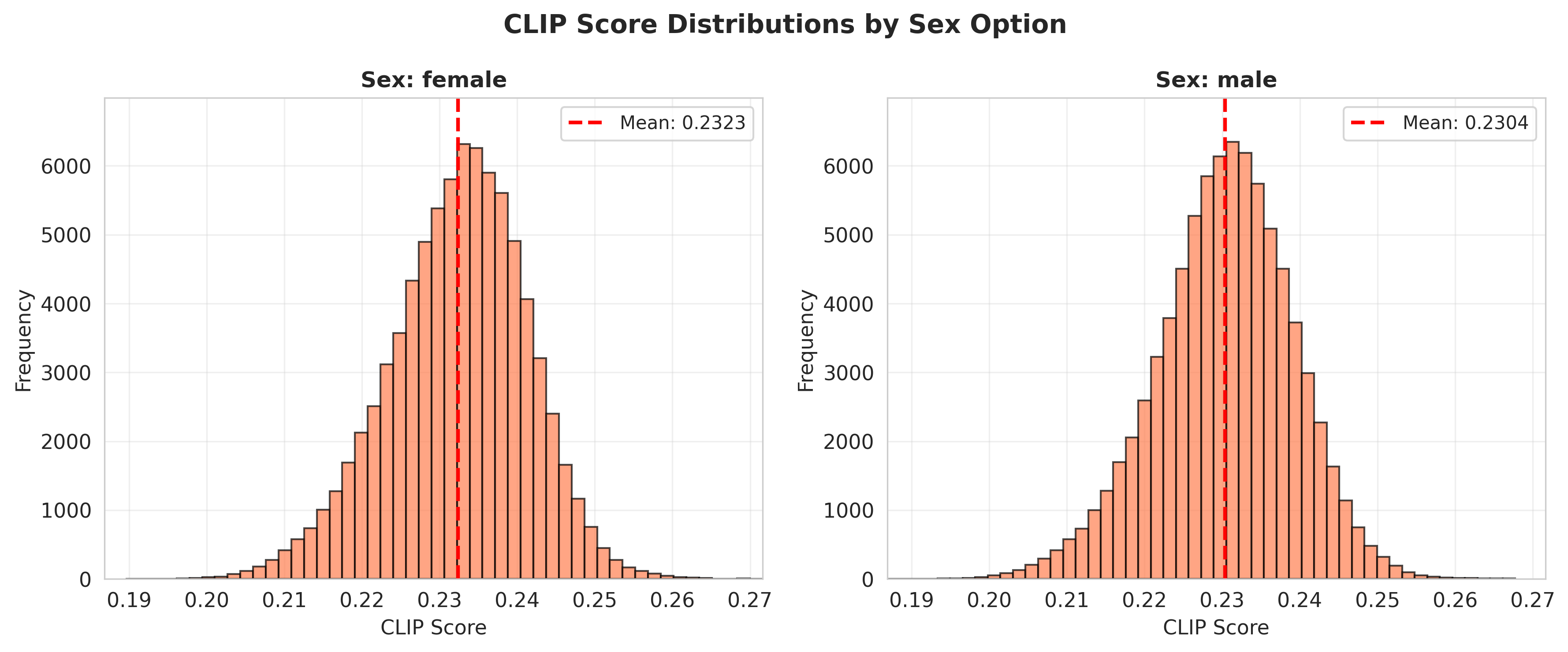}
    \caption{CLIP Score Distributions by Sex (synthetic)}
\end{figure*}

{
    \small
    \bibliographystyle{ieeenat_fullname}
    \bibliography{main}
}